%% file: cmas_arxiv.tex
\pdfoutput=1
\documentclass[10pt,twocolumn,letterpaper]{article}

\usepackage[pagenumbers]{cvpr} %

\input{preamble}
\usepackage{booktabs}
\usepackage{multirow}
\usepackage{amsmath,amssymb,bm}
\usepackage{xspace}
\usepackage{float}
\usepackage{placeins}
\usepackage{array}
\usepackage{tikz}
\usetikzlibrary{arrows.meta,positioning}

\definecolor{cvprblue}{rgb}{0.21,0.49,0.74}
\usepackage[pagebackref,breaklinks,colorlinks,allcolors=cvprblue]{hyperref}

\def\confName{CVPR}
\def\confYear{2027}

\newcommand{\ours}{CMAS\xspace}

\newcommand{\up}{$^{\uparrow}$}
\newcommand{\dn}{$^{\downarrow}$}

\definecolor{posgreen}{HTML}{168A55}
\definecolor{negred}{HTML}{A65C5C}
\newcommand{\pos}[1]{\textcolor{posgreen}{#1}}
\newcommand{\redneg}[1]{\textcolor{negred}{#1}}

\newcommand{\paran}[1]{\noindent \textbf{#1}.}
\newcommand{\para}[1]{\vspace{0.1cm} \noindent \textbf{#1}.}

\newcommand{\squishlist}{
    \begin{list}{$\bullet$}{ 
        \setlength{\itemsep}{0pt}
        \setlength{\parsep}{1pt}
        \setlength{\topsep}{1pt}
        \setlength{\partopsep}{0pt}
        \setlength{\leftmargin}{1.5em}
        \setlength{\labelwidth}{1em}
        \setlength{\labelsep}{0.5em} 
    } 
}
\newcommand{\squishend}{
  \end{list}  }

\title{Steering to the Point: Cross-Modal Activation Steering\\for Few-Shot Adaptation of Promptable Segmentation Models}

\author{Nuren Zhaksylyk \qquad Philipp Christmann \qquad Mario Fritz\\
CISPA Helmholtz Center for Information Security\\
{\tt\small \{nuren.zhaksylyk, philipp.christmann, fritz\}@cispa.de}\\[2pt]
{\small Project page: \url{https://zhnuren.github.io/cmas-project-page/}}
}

\providecommand{\PC}[1]{}
\begin{document}
\maketitle

\input{sec/0_abstract}

\input{sec/1_intro}

\input{sec/2_related}

\input{sec/3_diagnosis}

\input{sec/4_method}

\input{sec/5_experiments}

\input{sec/6_conclusion}

\section*{Acknowledgements}
This work was supported by the European Union's Horizon Europe research and innovation programme under grant agreement number 101214398 (ELLIOT).
Funded by the European Union. Views and opinions expressed are however those of the author(s) only and do not necessarily reflect those of the European Union or the European Commission. Neither the European Union nor the European Commission can be held responsible for them.

{
    \small
    \bibliographystyle{ieeenat_fullname}
    \bibliography{refs}
}

\input{sec/X_suppl}

\end{document}

%% file: sec/0_abstract.tex
\begin{abstract}
Promptable segmentation models
can identify and delineate objects from text descriptions alone.
However, even domain-adapted models
remain vulnerable to seemingly minor distribution shifts (e.g., different imaging protocols or target structures).
We first investigate segmentation errors
under such distribution shifts and find
that they are often subtle,
arising from calibration errors or slight misinterpretations of text prompts.
Our analysis shows that, in many cases,
a single click is sufficient to correct these errors,
demonstrating how \textit{cross-modal guidance} can overcome limitations of text-only prompting.
Based on these observations,
we introduce \textbf{C}ross-\textbf{M}odal \textbf{A}ctivation \textbf{S}teering (\ours)
for few-shot adaptation of promptable segmentation models.
Rather than introducing the cost and burden of additional expert clicks at inference time,
we derive \textit{cross-modal steering vectors} from
activation differences induced by adding a click to the text prompt.
The inference itself remains text-only, and applies the derived steering vectors to the model's
activations.
With only five annotated images,
\ours improves performance by up to 27.1 Dice points
across six datasets compared to text-only inference.
We
demonstrate that cross-modal steering generalizes across
datasets and
models,
supports bidirectional transfer between text and point prompts, and enables test-time adaptation.
Together, these results establish cross-modal steering as a
general paradigm for adapting promptable segmentation models without parameter updates.
\end{abstract}

%% file: sec/1_intro.tex
\section{Introduction}
\label{sec:intro}
\begin{figure}[t]
\centering
\includegraphics[width=\linewidth, trim=0 4pt 0 0, clip]{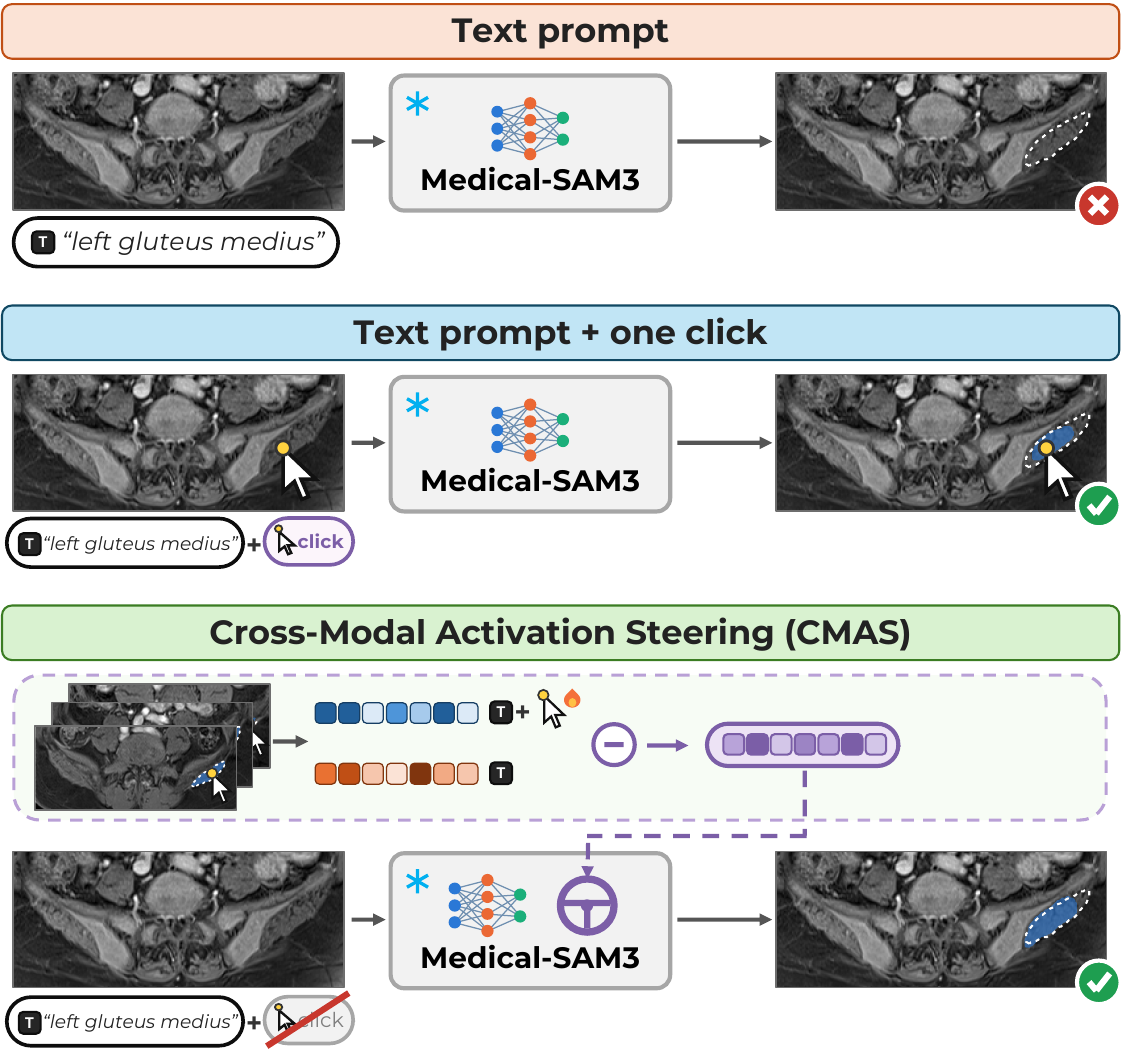}
\caption{Overview of \textit{Cross-Modal Activation Steering}
(CMAS), providing the
guidance of a click~\raisebox{-0.25ex}{\includegraphics[height=1.1em]{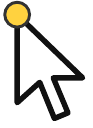}} for text-only inference.}
\label{fig:teaser}
\end{figure}

\paran{Motivation}
Recent promptable segmentation models
have substantially expanded the scope of image segmentation~\cite{carion2025sam3}.
Rather than requiring a dedicated model for each
segmentation
task,
they can identify and segment arbitrary concepts specified through
natural-language prompts.
Domain-specific adaptations further extend these capabilities
to specialized applications.
For instance, Medical-SAM3~\cite{jiang2026medicalsam3,liu2025medsam3}
enables
segmentation of anatomical structures
in medical scans, reducing the need for task-specific models.

Despite these advances, strong segmentation performance does not
necessarily transfer to new datasets, even within the same domain.
Even for domain-adapted models such as Medical-SAM3,
seemingly minor changes in acquisition 
conditions or target structures, such 
as different imaging protocols or segmenting teeth instead of organs, 
can be challenging.
General-purpose models face similar challenges in specialized visual settings,
including camouflaged animals, rare object categories, and aerial imagery.

\para{Prior work}
A natural response to these failures is further model adaptation,
for instance through parameter-efficient fine-tuning
such as LoRA~\cite{hu2022lora}.
However, collecting sufficient annotations is
challenging in domains requiring expert knowledge.
In practice, only a handful of annotated examples may be available,
making conventional adaptation prone to overfitting.
Alternative approaches based on prompt learning or confidence re-scoring avoid extensive model fine-tuning, but do not consistently address the underlying segmentation failures.

Crucially, these failures do not always reflect a lack of
segmentation capabilities.
Our failure-mode analysis reveals that errors frequently concern
object presence, semantic identification, or mask delineation.
We further observe that providing a single click inside the
requested structure often substantially improves the prediction.
This raises the key research question:
\textit{Can the corrective information provided by cross-modal guidance
be exploited without requiring such guidance at inference time?}

\para{Our approach}
We address this question through
\textit{Cross-Modal Activation Steering} (\textbf{\ours}),
which derives steering vectors from activation differences
induced by additional cross-modal guidance.
Activation steering modifies a model's behavior through
interventions on internal representations~\cite{
turner2023activation,li2023iti,zou2023repe,rimsky2024caa}.
While predominantly studied in language models,
we extend this principle to promptable segmentation
and introduce a cross-modal variation:
steering vectors are derived from activation differences
induced by additional prompt modalities.

Our key idea is to extract the representational changes
induced by corrective spatial prompts and reuse them
during inference with text prompts alone.
Given a small set of annotated examples,
we optimize point-prompt embeddings to improve segmentation.
We then compute activation differences between inputs
with and without the optimized point prompts.
These differences are averaged into class-specific steering vectors.
The intervention site and strength are selected automatically
using leave-one-out validation.
At inference time, the steering vectors are added to the
frozen model's activations, requiring neither additional clicks
nor parameter updates.
This allows us to \textit{steer to the point},
without requiring the point itself.

\para{Benefits of cross-modal steering}
We evaluate \ours across challenging segmentation benchmarks
spanning medical imaging, camouflage, long-tail objects, and aerial imagery.
With only five annotated images, \ours improves segmentation
performance by 5.0--27.1 Dice points across six datasets,
compared to text-only prompting.
Moreover, \ours provides complementary gains when combined with existing adaptation approaches, and provides fewer false positives than 
baselines when the target class is absent.

More broadly, cross-modal steering introduces a way to
exploit interactions between input modalities without requiring all modalities at inference time.
While we instantiate this principle for promptable segmentation, its formulation is independent of a particular segmentation architecture or prompt modality.
Our experiments demonstrate that the idea generalizes across datasets and 
models. We also find that cross-modal steering helps in the reverse direction
with textual cross-modal guidance distilled into steering vectors for click-only inference, and that it also enables test-time adaptation.
Finally, \ours enables learning from clicks-only, without access
to ground-truth segmentation masks.

\para{Contributions}
\begin{itemize}
\item Conducting an in-depth failure-mode analysis of promptable
segmenters under distribution shifts, revealing the corrective
potential of cross-modal guidance.
\item Introducing cross-modal steering as a general approach
for transferring information between input modalities
through activation interventions.
\item Proposing \ours for low-resource segmentation
via cross-modal steering, with lightweight prompt optimization
and automated steering site and strength selection.
\item Demonstrating the effectiveness of \ours
across datasets, models, and prompt modalities,
and releasing our framework to facilitate adoption and reproducibility.
\end{itemize}

%% file: sec/2_related.tex
\section{Related Work}
\label{sec:related}

\paran{Promptable segmentation}
Foundation segmentation models enable segmentation
of arbitrary objects through spatial prompts,
such as points, boxes, or masks.
SAM~\cite{kirillov2023sam} and SAM2~\cite{ravi2024sam2}
established strong general-purpose segmentation capabilities,
while models such as SEEM~\cite{zou2023seem}, GLEE~\cite{gleewu2024general}
and SAM3~\cite{carion2025sam3} additionally support
textual prompts.
Domain-specific adaptations further extend these capabilities
to specialized settings.
One particular application is medical imaging,
with specialized models
through full training~\cite{zhao2025sat,zhao2025biomedparse} or fine-tuning~\cite{ma2024medsam,ma2025medsam2,
wang2023sammed3d,jiang2026medicalsam3,liu2025medsam3}
or lightweight adapters~\cite{wu2023medsa,chen2023masam}.
Despite these advances, promptable segmentation models
can struggle with unfamiliar structures and visual conditions,
motivating adaptation with limited annotations.

\para{Few-shot adaptation}
Existing approaches adapt pretrained segmentation models
through several mechanisms.
Parameter-efficient fine-tuning updates selected model
parameters~\cite{hu2022lora,xiao2024catsam,
nagaraju2026fewcases}, while prompt learning optimises
input embeddings with frozen model weights
~\cite{zhou2022coop,ptsam2025,yan2025pgpsam,
venkataramani2026fscpl,zhou2026m2c}.
Reference-based approaches exploit annotated examples
through masks~\cite{zhang2024persam,liu2024matcher,
zhang2024gfsam,protosam2024}, visual prompts~\cite{sun2024vrpsam,
wang2023seggpt}, or in-context segmentation
~\cite{butoi2023universeg,rakic2024tyche}.
Other methods adjust prediction scores through
calibration~\cite{guo2017calibration,kuzucu2024detcalib}
or class prototypes~\cite{rosi2026prada}.
In contrast, \ours adapts internal activations
without updating model weights or requiring
additional inference prompts.

\para{Activation steering}
Activation steering modifies model behavior through
interventions on internal representations, often
using activation differences between contrasting inputs,
and has been primarily 
explored in language models
~\cite{turner2023activation,li2023iti,zou2023repe,
rimsky2024caa}.
Related work investigates task vectors derived from in-context examples
~\cite{todd2024functionvectors,liu2024icv},
learned representation interventions~\cite{wu2024reft},
and steering for low-resource adaptation
~\cite{weights2activations2026}.
In vision, activation interventions have been used
to reduce hallucinations in vision-language models~\cite{liu2025vti},
while visual task vectors~\cite{hojel2024visualtaskvectors} capture
task information from in-context examples.
We extend activation steering to promptable
segmentation models, deriving interventions from
the representational changes induced by additional
prompt modalities.

\para{Cross-modal transfer}
Multimodal prompting combines complementary inputs,
such as text and spatial prompts~\cite{zou2023seem,carion2025sam3}.
Related work has shown that task representations
can transfer across modalities through activation
interventions in vision-language models
~\cite{luo2025crossmodaltask}.
\ours distills cross-modal guidance into steering vectors,
enabling inference without additional modalities or model updates.

%% file: sec/3_diagnosis.tex
\section{Failure-Mode Analysis}
\label{sec:diagnosis}

Despite their broad capabilities,
promptable segmentation models may struggle when applied to tasks
that differ from their training distribution.
Such shifts can arise from changes in acquisition conditions,
target structures, object appearance, or class frequencies.
This section investigates how these seemingly subtle shifts
affect segmentation
and which failure modes arise.
We then test wether cross-modal guidance can correct these errors.

\subsection{Datasets}

We select six challenging datasets representing different
types of distribution shifts, including settings requiring
specialized knowledge or fine-grained visual understanding.

For medical imaging, we consider
\textbf{RAOS}~\cite{luo2024raos}, \textbf{ToothFairy2}~\cite{bolelli2026toothfairy2}, and
\textbf{MRISegmenter}~\cite{zhuang2025mrisegmenter}, which are outside the training
pool of the domain-specific segmenter Medical-SAM3~\cite{jiang2026medicalsam3}.
RAOS introduces differences in anatomical coverage
and CT acquisition characteristics,
ToothFairy2 shifts to dental structures and cone-beam CT,
and MRISegmenter introduces different anatomical targets
and MRI acquisition conditions.

For the general-purpose model SAM3, we consider
\textbf{MoCA-Mask}~\cite{cheng2022mocamask}, featuring camouflaged animals
whose appearance closely resembles their surroundings;
\textbf{LVIS-92i}~\cite{gupta2019lvis}, targeting
rare and long-tailed object categories;
and \textbf{iSAID-5i}~\cite{zamir2019isaid},
featuring aerial imagery with small, densely arranged
objects and substantial scale variation.

Together, these datasets expose complementary challenges
in acquisition conditions, anatomical targets,
foreground-background similarity, class frequencies,
and object scale.
They allow us to examine how segmentation failures
vary across different forms of seemingly subtle distribution shift.

\subsection{Failure Modes}
Let $P$ and $G$ denote the predicted and ground-truth masks
for a given prompt,
and $O$
the pixels of
all \emph{other} annotated
structures in the image or slice.
We distinguish five failure modes:
\squishlist
\item \textbf{Presence false-positive}:
$P\neq\emptyset$, $G=\emptyset$;
the model predicts a structure although the target is absent.
\item \textbf{Presence false-negative}:
$P=\emptyset$, $G\neq\emptyset$;
the model misses a target that is present.
\item \textbf{Semantic mismatch}: $(P\setminus G)\cap O\neq\emptyset$;
  predicted pixels that belong to a different annotated structure
  than requested (e.g., the left instead of the right
  femoral head).
\item \textbf{Mask false-positive area}: $(P\setminus G)\setminus O\neq\emptyset$;
the prediction extends into unannotated regions.
\item \textbf{Mask false-negative area}: $G\setminus P\neq\emptyset$;
the prediction omits regions of the target.
\squishend

\noindent As these failure modes may overlap, we employ hierarchical
attribution to decompose the Dice deficit.
Presence errors take precedence over semantic mismatches,
which in turn take precedence over mask delineation errors.
Let $E_k$ denote the erroneous pixels attributed to failure
mode $k$, such that $E_i\cap {E}_j=\emptyset$ for $i\neq j$
and $\bigcup_{k=1}^{5}E_k=P\triangle G$,
where $\triangle$ denotes the symmetric difference.
This yields
\begin{equation}
\underbrace{\frac{2|P\cap G|}{|P|+|G|}}_{\mathrm{Dice}}
+
\sum_{k=1}^{5}
\underbrace{\frac{|E_k|}{|P|+|G|}}_{w_k}
=1,
\end{equation}
where $w_k$ quantifies the contribution of failure mode $k$
to the Dice deficit.
We average
$w_k$
across predictions.

\begin{table}[t]
\centering
\small
\setlength{\tabcolsep}{1.5pt}
\resizebox{\linewidth}{!}{%
\begin{tabular}{@{}lcccccc@{}}
\toprule
 & \multicolumn{3}{c}{\textbf{Medical} (Medical-SAM3)} & \textbf{Camouflage} & \textbf{Long-tail} & \textbf{Aerial} \\
\cmidrule(lr){2-4}\cmidrule(lr){5-5}\cmidrule(lr){6-6}\cmidrule(lr){7-7}
 & RAOS & ToothFairy2 & MRISeg. & MoCA-Mask & LVIS-92i & iSAID-5i \\
\midrule
Text prompt  & 0.553 & 0.016 & 0.463 & 0.351 & 0.545 & 0.389 \\
\midrule
\multicolumn{7}{@{}l}{\textit{Presence}} \\
Presence FN & +0.045 & +0.419 & +0.085 & +0.573 & +0.352 & +0.447 \\
Presence FP & +0.180 & +0.404 & +0.248 & NA & NA & NA \\
\midrule
\multicolumn{7}{@{}l}{\textit{Confusion}} \\
Semantic mismatch & +0.102 & +0.036 & +0.109 & NA & +0.019 & +0.040 \\
\midrule
\multicolumn{7}{@{}l}{\textit{Mask quality}} \\
FN area & +0.105 & +0.080 & +0.064 & +0.022 & +0.049 & +0.066 \\
FP area  & +0.013 & +0.045 & +0.029 & +0.053 & +0.036 & +0.058 \\

\midrule
\textbf{Cross-modal guidance}  & 0.595{\scriptsize$\pm$.013} & 0.202{\scriptsize$\pm$.012} & 0.562{\scriptsize$\pm$.015} & 0.636{\scriptsize$\pm$.059} & 0.771{\scriptsize$\pm$.049} & 0.568{\scriptsize$\pm$.079} \\

\bottomrule
\end{tabular}}
\caption{Dice decomposition into five failure modes.
FN/FP: false-negative/positive; NA: not applicable.
Cross-modal guidance is averaged over three random clicks (with $\pm$ standard deviation).}
\label{tab:errors}
\end{table}

\para{Findings}
\Cref{tab:errors} decomposes the Dice deficit into
individual failure modes.
We make two key observations.

First, {presence errors account for a substantial
fraction of the performance loss}.
Missed structures alone contribute 0.419 Dice points
on ToothFairy2, 0.573 on MoCA-Mask, and 0.447 on iSAID-5i.
On the medical datasets, false-positive presence predictions
further contribute between 0.180 and 0.404 Dice points.
This shows that models often fail to even detect the requested structures.

Second, {semantic confusion and mask-quality errors
also contribute considerably}.
On RAOS and MRISegmenter, semantic mismatches account
for 0.102 and 0.109 Dice points, respectively,
while mask-quality errors contribute another
0.118 and 0.093 Dice points.
Thus, even when structures are detected, models may
identify the wrong target or produce inaccurate masks.

\para{Implications for adaptation}
The diversity of failure modes poses challenges for
adaptation approaches that only adjust confidence scores
or re-rank existing predictions.
While such methods can correct selection errors,
they cannot recover suitable masks that the model
fails to produce.

\subsection{Cross-Modal Guidance}

Our analysis indicates shortcomings of promptable segmentation models
under seemingly minor distribution shifts. 
A common strategy for improving text-prompted segmentation
is to provide additional inputs, such as expert clicks,
which offer cross-modal guidance.

\para{Simulating expert clicks}
As actual expert clicks are unavailable, we simulate them
by randomly sampling a positive point from the ground-truth mask.
While these simulated clicks may not reflect actual expert interactions,
they provide a proxy for evaluating cross-modal guidance.
To account for variability in click placement,
we repeat the experiment three times and report
the mean Dice score and standard deviation.
We evaluate the same six datasets as before,
comparing against text-only prompting.

\para{Cross-modal guidance recovers errors}
The last row of \Cref{tab:errors} reports the results.
Adding a single click substantially improves segmentation performance
on all six dataset, even though this click
is randomly sampled from the ground-truth mask
and potentially suboptimal.
For example, performance
on MoCA-Mask improves from 0.351 to 0.636,
on LVIS-92i from 0.545 to 0.771,
and on iSAID-5i from 0.389 to 0.568.
These results demonstrate that cross-modal guidance,
even through a single click,
can substantially improve segmentation across diverse settings.

%% file: sec/4_method.tex
\section{Cross-Modal Activation Steering (CMAS)}
\label{sec:method}

Our failure-mode analysis in \Cref{sec:diagnosis} reveals
that text-only prompting frequently fails at identifying
or delineating the requested structures.
Yet, providing cross-modal guidance via a single additional click often corrects
these errors, suggesting that the model possesses segmentation
capabilities that text-only prompting fails to elicit.

\begin{figure*}[t]
\centering
\includegraphics[width=0.9\textwidth, trim=0 0 0 33pt, clip]{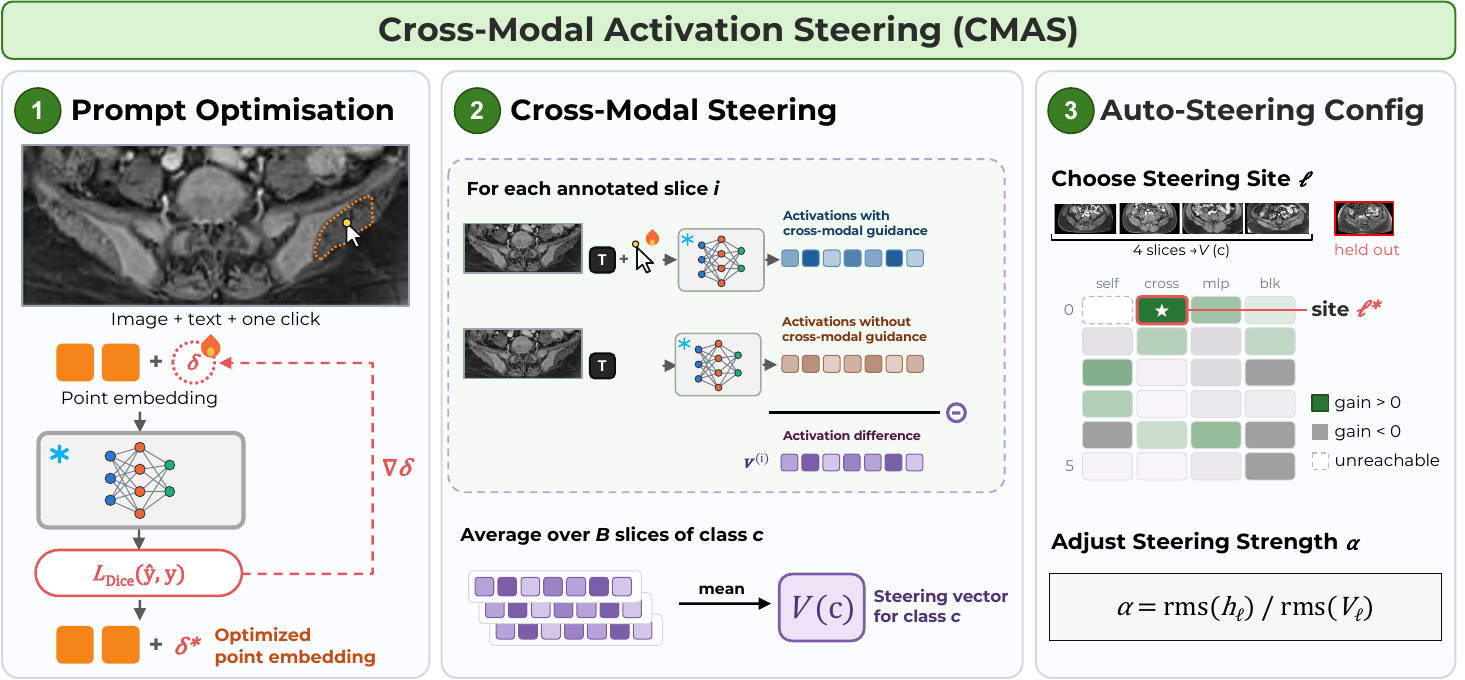}
\caption{\textbf{Overview of \ours.}
(1) Prompt embeddings are optimised using annotated examples.
(2) Cross-modal steering vectors are derived from activation
differences with and without additional modality guidance,
and averaged across examples.
(3) The steering site and strength are configured automatically.
At inference, the resulting vectors steer the frozen model
without requiring the additional modality.}
\label{fig:method}
\end{figure*}

Motivated by this observation, we introduce \textit{Cross-Modal Activation Steering} (\ours),
which transfers the corrective effect of spatial guidance
to text-only prompting.
Using only a few annotated examples, we extract
cross-modal steering vectors from activation changes
induced by corrective clicks (\Cref{sec:method_steer}).
These vectors are applied to the frozen model,
without requiring spatial guidance at inference time.
In addition, we develop a mechanism for optimising point
embeddings to account for potentially suboptimal points (\Cref{sec:method_click}).
Our steering intervention is configured automatically to avoid
costly and cumbersome tuning (\Cref{sec:method_config}).
\Cref{sec:method_notation} introduces our notation.
An overview of \ours is provided in \Cref{fig:method}.

\subsection{Notation}
\label{sec:method_notation}

\paran{Modalities}
Let $x=\{m_1,\dots,m_K\}$ denote an input consisting of
$K$ modalities available at inference.
For example, $m_1$ could be an image and $m_2$ a text prompt.
Each modality $m_j$ is processed by a modality-specific
encoder $\mathcal{E}_j$, yielding the embedding $\mathcal{E}_j(m_j)$.
The promptable segmenter model $M$ predicts
$y=M(x)=M(\{m_1,\dots,m_K\})$,
from which the final segmentation mask $P$ is obtained.

\para{Sites}
For an intervention site $\ell$, such as the cross-attention
of a fusion-encoder layer, 
let $H_\ell$ denote the model
computation up to this site, yielding the activation
$h_\ell=H_\ell(x)$.
We denote the remaining model computation by $F_\ell$,
such that $y=M(x)=F_\ell(h_\ell)$.

\subsection{Cross-Modal Steering Vectors}
\label{sec:method_steer}

Our central idea is to exploit an additional input modality
to derive steering vectors, without requiring that modality
at inference time.
Formally, let $x_+=x\cup\{m_+\}$ denote the input $x$
augmented with an additional modality $m_+$.
For example, for text-prompted segmentation, $x$ contains the image
and text prompt, while $m_+$ may represent a click.

\para{Derivation of steering vectors}
We derive cross-modal steering vectors from activation
differences between inputs with and without additional
modality guidance.
For an adaptation example $x^{(i)}$, we obtain:
\begin{equation}
v_\ell^{(i)}
=
H_\ell(x_+^{(i)})-H_\ell(x^{(i)})
\label{eq:cms}
\end{equation}
This difference captures the representational changes
induced by the additional modality, which we aim to
transfer to inference without it.
This is different from conventional intra-modal steering
that typically contrasts
activations from inputs yielding positive and negative
model outcomes,
given the same input modalities:
$H_\ell(x^{P})-H_\ell(x^{N})$.
In contrast, cross-modal steering derives activation differences
from inputs with and without cross-modal guidance.

Given $B$ adaptation examples,
we average their activation differences into
a steering vector for class $c$:
\begin{equation}
V_\ell(c)
=
\frac{1}{B}\sum_{i=1}^{B}v_\ell^{(i)}.
\label{eq:steering_vector}
\end{equation}
As activation differences may capture
instance-specific spatial information,
we aggregate them across examples of the same class
to obtain usable steering vectors.

\para{Steering at inference}
At test time, the input $x$ is restricted to modalities in $K$, without guidance through $x_+$.
We apply the steering vector at site $\ell$:
\begin{equation}
h_\ell^\ast=H_\ell(x)+\alpha\,V_\ell(c),
\label{eq:steer}
\end{equation}
where $\alpha$ controls the steering strength.
$\ell$ and $\alpha$ are chosen automatically (\Cref{sec:method_config}).
The prediction is then obtained from the remaining
model computation:
\begin{equation}
y^\ast=F_\ell\big(h_\ell^\ast\big)
\label{eq:steering-inference}
\end{equation}
The final segmentation mask is obtained from $y^\ast$
as before.

At image-token sites, $V_\ell(c)\in\mathbb{R}^{T\times d}$
contains one $d$-dimensional vector per image token.
Rather than applying the same direction at every position,
the steering vector preserves spatial variations in
the activation changes induced by cross-modal guidance.

\subsection{Prompt Optimization}
\label{sec:method_click}

The effectiveness of cross-modal steering depends on
the quality of the guidance used to derive steering vectors.
However, expert clicks are rarely annotated, and
randomly sampling point from the ground-truth mask
can lead to suboptimal guidance.
This can limit the corrective information
captured in the steering vectors.
We therefore optimise the prompt embeddings of the additional modality
to strengthen their cross-modal guidance during steering.

Let $\delta$ denote a continuous perturbation applied
exclusively to this modality's embedding,
such that $x+\delta$ replaces $\mathcal{E}_j(m_j)$ with
$\mathcal{E}_j(m_j)+\delta$.
All other embeddings are left unchanged.
Given a target mask $G$ and task loss $\mathcal{D}$,
we obtain
\begin{equation}
\delta^\ast
=
\operatorname*{arg\,min}_{\delta\in\mathbb{P}}
\mathcal{D}\big(M(x+\delta),G\big).
\label{eq:prompt-opt}
\end{equation}
The optimized cross-modal steering direction becomes
\begin{equation}
v_\ell^{(i)}
=
H_\ell(x_+^{(i)}+\delta^{(i)\ast})
-
H_\ell(x^{(i)}).
\label{eq:vc-optimized}
\end{equation}

\para{Point optimization}
For each annotated mask, we randomly sample a positive
click from the target region
and optimise its point-token and geometry-CLS embeddings.
Note that the optimised embeddings need not correspond to
a valid point in the original input space.
We use Adam for 40 steps with a learning rate of 0.1,
minimising the soft Dice loss between the merged
prediction and the ground-truth mask, and retain
the best optimisation step.
The model weights remain frozen, and the per-example
perturbation is discarded after extracting the activations.
We evaluate the benefit over unmodified clicks in
\Cref{sec:experiments}.

\para{Generality}
While we optimise point prompts in our main experiments,
our formulation permits optimisation of other modalities as well.
We also investigate the reverse direction of cross-modal
steering, using text guidance for point-only inference
in our experiments.

\subsection{Auto-Steering Configuration}
\label{sec:method_config}

The effectiveness of activation steering depends critically
on the intervention site and steering strength.
However, suitable configurations can vary substantially
across models and datasets, making manual tuning
cumbersome and computationally expensive.
This poses a challenge, especially in few-shot settings
where no separate validation set is available.
We therefore introduce an approach for automatic configuration of the steering,
which determines both parameters
using only the available adaptation examples.

\para{Site}
Different intervention sites capture different aspects
of the model's internal representations, and not all
are equally suitable for steering.
We therefore select suitable sites through
leave-one-out validation on the annotated examples.
Given an adaptation budget $B$, for each candidate site $\ell$ and held-out example,
we construct class vectors from the remaining
$B-1$ examples and evaluate steered predictions
on the held-out example.
We select the site with the highest mean Dice gain
relative to the text-only baseline
across held-out examples and classes, applying steering
only when this gain is positive.
The selected site is then used for test inference
(\cref{eq:steer}).
For SAM3, we consider 54 candidate sites across the
fusion encoder and decoder, including individual layers
and their attention and MLP sub-modules.

\para{Strength}
The effectiveness of steering also depends on the strength
$\alpha$ of the intervention.
However, activation magnitudes can vary substantially
across sites and models, making a fixed strength
unreliable and manual tuning cumbersome.
Searching for suitable strengths at every candidate site
would further increase the configuration cost.
We instead compute $\alpha$
by matching
the magnitude of the steering intervention to that of
the original activations:
\begin{equation}
\alpha
=
\frac{\mathrm{rms}(h_\ell)}
     {\mathrm{rms}(V_\ell)},
\label{eq:strength}
\end{equation}
where $\mathrm{rms}(h_\ell)$ denotes the mean RMS (root mean square)
of text-only activations over annotated examples,
and $\mathrm{rms}(V_\ell)$ the mean RMS of the
class-specific steering vectors at site $\ell$.
This provides scale-aware steering across sites
and models without additional hyperparameter tuning.

\subsection{Computational Costs}

Constructing the steering vectors requires one text-only
forward pass, 40 forward--backward passes for prompt
optimisation (one for each optimization step),
and one forward pass with the optimised
prompt per annotated example.
Auto-Steering requires one steered forward
pass per held-out example and candidate site.
At inference, \ours introduces only a single activation
addition, with negligible computational overhead
relative to text-only inference.

%% file: sec/5_experiments.tex
\section{Experiments}
\label{sec:experiments}

\subsection{Setup}

\paran{Datasets}
We use the same six datasets as in \Cref{sec:diagnosis}: 
\textbf{RAOS}~\cite{luo2024raos}, \textbf{ToothFairy2}~\cite{bolelli2026toothfairy2},
\textbf{MRISegmenter}~\cite{zhuang2025mrisegmenter}, 
from the medical domain,
and  \textbf{MoCA-Mask}~\cite{cheng2022mocamask} (camouflaged animals)
\textbf{LVIS-92i}~\cite{gupta2019lvis} folds 0-2 (rare and long-tail object categories)
and \textbf{iSAID-5i}~\cite{zamir2019isaid} (aerial images).

\para{Budget}
For \ours and all baselines, the adaptation budget is restricted to five slices per dataset-class, randomly drawn from the training split (seed 42).
For volumes in the medical domain, only the middle slice is annotated.

\para{Segmentation models}
For the medical datasets, we use Medical-SAM3 V2%
\footnote{\href{https://github.com/AIM-Research-Lab/Medical-SAM3}{AIM-Research-Lab/Medical-SAM3}}
and SAM3%
\footnote{\href{https://github.com/facebookresearch/sam3}{facebookresearch/sam3}}
for all others.

\para{Baselines}
We compare \ours against representative few-shot
adaptation methods with the same budget $B$.
These include prompt learning, reference-based
adaptation, parameter-efficient fine-tuning,
and prediction re-scoring.
We additionally evaluate intra-modal steering
and combinations of \ours with LoRA and PrAda.

\para{Metrics}
We report Dice as the primary metric, and
Normalized Surface Dice (NSD)
in the supplement.
Statistical significance is assessed using paired
Wilcoxon signed-rank tests.
\up/\dn indicate improvements/degradations
relative to text-only prompting ($p<0.05$).

\begin{table}[t]
\centering
\small
\setlength{\tabcolsep}{1.5pt}
\resizebox{\linewidth}{!}{%
\begin{tabular}{@{}lcccccc@{}}
\toprule
 & \multicolumn{3}{c}{\textbf{Medical} (Medical-SAM3)}
 & \textbf{Camouflage} & \textbf{Long-tail} & \textbf{Aerial} \\
\cmidrule(lr){2-4}\cmidrule(lr){5-5}
\cmidrule(lr){6-6}\cmidrule(lr){7-7}
 & RAOS & ToothFairy2 & MRISeg.
 & MoCA-Mask & LVIS-92i & iSAID-5i \\
\midrule
\multicolumn{7}{@{}l}{\textit{Text-only inference}} \\
Default
 & 0.553 & 0.016 & 0.463 & 0.351 & 0.545 & 0.389 \\
\textbf{\ours (click $\rightarrow$ text)}
 & \textbf{0.642} & \textbf{0.132} & \textbf{0.513}
 & \textbf{0.622} & \textbf{0.657} & \textbf{0.523} \\
$\Delta$Dice (\ours)
 & \pos{+0.089} & \pos{+0.116} & \pos{+0.050}
 & \pos{+0.271} & \pos{+0.112} & \pos{+0.134} \\
\midrule
\multicolumn{7}{@{}l}{\textit{Click-only inference}} \\
Default
 & 0.692 & 0.375 & 0.640 & \textbf{0.659} & 0.679 & 0.562 \\
\textbf{\ours (text $\rightarrow$ click)}
 & \textbf{0.745} & \textbf{0.484} & \textbf{0.680}
 & 0.592 & \textbf{0.697} & \textbf{0.593} \\
$\Delta$Dice (\ours)
 & \pos{+0.053} & \pos{+0.109} & \pos{+0.040}
 & \redneg{-0.067} & \pos{+0.018} & \pos{+0.031} \\
\bottomrule
\end{tabular}}
\caption{Benefits of cross-modal steering for text-only and
click-only inference, using five annotated images per class.
$\Delta$Dice denotes improvement of \ours over text-only/click-only inference.}
\label{tab:steering-benefits}
\end{table}

\subsection{Benefits of Cross-Modal Steering}

We investigate whether cross-modal steering provides a general mechanism for adapting promptable segmentation models.
Beyond text-only inference through click guidance, we examine its effectiveness in the reverse direction, across model architectures, and under different supervision
settings, including test-time adaptation.

\para{Improving text-only segmentation}
\Cref{tab:steering-benefits} compares cross-modal steering from clicks against text-only inference.
\ours improves performance across all six datasets by 5.0--27.1 Dice points, with the largest gains
on MoCA-Mask (+27.1), iSAID-5i (+13.4),
and ToothFairy2 (+11.6).
These results demonstrate that cross-modal steering
substantially improves promptable segmentation,
without updating any model parameters.
Notably, this improvement is achieved with
only five annotated images per class.

\para{Generalization across prompt modalities}
We also investigate the reverse direction,
demonstrating the versatility of cross-modal steering.
Specifically, we derive steering vectors from text guidance
and apply them during click-only inference.
As shown in \Cref{tab:steering-benefits},
this also improves performance on five of six datasets,
including ToothFairy2 (+10.9)
and RAOS (+5.3).
Only MoCA-Mask exhibits a performance decrease (-6.7).
Upon closer investigation, we find that even oracle cross-modal guidance does not help on this dataset (details are provided in \Cref{sup:clicktransfer}).

\para{Adaptation from clicks alone}
Unlike conventional few-shot adaptation methods,
\ours can derive steering vectors from clicks alone,
without requiring ground-truth masks.
\Cref{tab:no-masks} evaluates this setting with
1--5 clicks per support image.
Even with a single click, \ours improves performance
on five of six datasets, with gains of up to 19.8 Dice points.
Performance further improves for 3 and 5 clicks (except for MRISeg.).
Direct cross-modal guidance at inference serves as
an oracle reference for the information available from this additional modality.
Depending on the dataset, \ours recovers up to 83\%
of the gains achieved by oracle guidance,
while retaining text-only inference.

\begin{table}[t]
\centering
\small
\setlength{\tabcolsep}{1.5pt}
\resizebox{\linewidth}{!}{%
\begin{tabular}{@{}lcccccc@{}}
\toprule
 & \multicolumn{3}{c}{\textbf{Medical} (Medical-SAM3)}
 & \textbf{Camouflage} & \textbf{Long-tail} & \textbf{Aerial} \\
\cmidrule(lr){2-4}\cmidrule(lr){5-5}
\cmidrule(lr){6-6}\cmidrule(lr){7-7}
 & RAOS & ToothFairy2 & MRISeg.
 & MoCA-Mask & LVIS-92i & iSAID-5i \\
\midrule
Text-only prompt
 & 0.553 & 0.016 & 0.463 & 0.351 & 0.545 & 0.389 \\
\midrule
\multicolumn{7}{@{}l}{\textit{\ours: Text-only inference}} \\
1 click/image
 & 0.552{\scriptsize$\pm$.005}
 & 0.063{\scriptsize$\pm$.027}
 & 0.486{\scriptsize$\pm$.006}
 & 0.549{\scriptsize$\pm$.009}
 & 0.676{\scriptsize$\pm$.001}
 & 0.537{\scriptsize$\pm$.005} \\
3 clicks/image
 & 0.552{\scriptsize$\pm$.006}
 & 0.096{\scriptsize$\pm$.031}
 & \textbf{0.488}{\scriptsize$\pm$.001}
 & 0.580{\scriptsize$\pm$.024}
 & 0.684{\scriptsize$\pm$.004}
 & 0.570{\scriptsize$\pm$.008} \\
5 clicks/image
 & \textbf{0.568}{\scriptsize$\pm$.002}
 & \textbf{0.120}{\scriptsize$\pm$.011}
 & 0.435{\scriptsize$\pm$.031}
 & \textbf{0.583}{\scriptsize$\pm$.007}
 & \textbf{0.686}{\scriptsize$\pm$.004}
 & \textbf{0.585}{\scriptsize$\pm$.003} \\
\midrule
\multicolumn{7}{@{}l}{\textit{Cross-modal guidance: Text + clicks at inference} (oracle guidance)} \\
1 click/image
 & 0.595{\scriptsize$\pm$.013}
 & 0.202{\scriptsize$\pm$.012}
 & 0.562{\scriptsize$\pm$.015}
 & 0.636{\scriptsize$\pm$.059}
 & 0.771{\scriptsize$\pm$.049}
 & 0.568{\scriptsize$\pm$.079} \\
3 clicks/image
 & 0.650{\scriptsize$\pm$.010}
 & 0.267{\scriptsize$\pm$.013}
 & 0.597{\scriptsize$\pm$.013}
 & 0.735{\scriptsize$\pm$.044}
 & 0.839{\scriptsize$\pm$.030}
 & 0.710{\scriptsize$\pm$.060} \\
5 clicks/image
 & 0.672{\scriptsize$\pm$.008}
 & 0.296{\scriptsize$\pm$.013}
 & 0.606{\scriptsize$\pm$.011}
 & 0.759{\scriptsize$\pm$.039}
 & 0.846{\scriptsize$\pm$.024}
 & 0.746{\scriptsize$\pm$.047} \\
\bottomrule
\end{tabular}}
\caption{Adaptation from clicks alone, without ground-truth masks.
\ours derives steering vectors from 1--5 clicks per support image
(five images per class), retaining text-only inference.
For reference, cross-modal guidance uses clicks directly
on test images. Results show mean Dice $\pm$ standard deviation
over three random click samples.}
\label{tab:no-masks}
\end{table}

\para{Generalization across models}
To assess whether cross-modal steering generalizes
beyond SAM3-based architectures, we evaluate \ours
with GLEE on MoCA-Mask, LVIS-92i and iSAID-5i.
\ours improves Dice from 0.167 to 0.287 on MoCA-Mask, from 0.544 to 0.576 on LVIS-92i, and from 0.122 to 0.394 on iSAID-5i,
demonstrating
that
cross-modal steering generalizes across models.

\para{Test-time adaptation}
At inference time, there might be no prior adaptation data
available at all.
In this experiment, we investigate whether cross-modal steering
can enable test-time adaptation for such cases.
Specifically, given three expert clicks on one slice/frame
of the given volume/video,
we derive a case-specific steering vector.
This vector is then applied to the remaining slices/frames.
This improves Dice over text-only prompting on all four
datasets with volumes/videos, including ToothFairy2 (0.016 to 0.201) and MoCA-Mask (0.351 to 0.573).
These results demonstrate that cross-modal steering
enables immediate expert guidance at inference
without updating model weights
(details are provided in \Cref{sup:tta} in the supplement).

\begin{table}[t]
\centering
\small
\setlength{\tabcolsep}{3pt}
\resizebox{\columnwidth}{!}{%
\begin{tabular}{@{}lcccccc@{}}
\toprule
 & \multicolumn{3}{c}{Medical-SAM3}
 & \multicolumn{3}{c}{SAM3} \\
\cmidrule(lr){2-4}
\cmidrule(lr){5-7}
Steering configuration
 & RAOS & TF2 & MRI & MoCA & LVIS & iSAID \\
\midrule
Text-only prompt (no steering)
 & 0.553 & 0.016 & 0.463 & 0.351 & 0.545 & 0.389 \\
\midrule
RAOS: \texttt{enc.0.cross} ($\alpha=1.68$)
 & * & 0.097 & \textbf{0.514} & 0.305 & 0.552 & 0.465 \\
TF2: \texttt{enc.1} ($\alpha=2.79$)
 & 0.591 & * & 0.481 & 0.248 & 0.367 & 0.325 \\
MRI: \texttt{enc.0.cross} ($\alpha=1.84$)
 & \textbf{0.643} & 0.103 & * & 0.302 & 0.549 & 0.466 \\
MoCA: \texttt{enc.4} ($\alpha=3.24$)
 & 0.553 & 0.023 & 0.429 & * & 0.617 & 0.443 \\
LVIS: \texttt{enc.3.self} ($\alpha=3.15$)
 & 0.629 & 0.033 & 0.489 & 0.186 & * & 0.519 \\
iSAID: \texttt{enc.1} ($\alpha=3.61$)
 & 0.560 & 0.123 & 0.439 & 0.243 & 0.282 & * \\
\midrule
\textbf{\ours + Auto-Steering}
 & \textbf{0.642} & \textbf{0.132} & \textbf{0.513}
 & \textbf{0.622} & \textbf{0.657} & \textbf{0.523} \\
\bottomrule
\end{tabular}}
\caption{Impact of our Auto-Steering Configuration.
Each row transfers the site and strength selected on
one dataset to the others (test Dice).
Diagonal entries (*) correspond to the Auto-Steering
results in the last row.
TF2: ToothFairy2; MRI: MRISegmenter; MoCA: MoCA-Mask.}

\label{tab:auto-steering}
\end{table}

\begin{table*}[t]
\centering
\small
\setlength{\tabcolsep}{5pt}
\resizebox{0.85\textwidth}{!}{%
\begin{tabular}{@{}llcccccc@{}}
\toprule
 & & \multicolumn{3}{c}{\textbf{Medical} (Medical-SAM3)}
     & \textbf{Camouflage} (SAM3)
     & \textbf{Long-tail} (SAM3)
     & \textbf{Aerial} (SAM3) \\
\cmidrule(lr){3-5}
\cmidrule(lr){6-6}
\cmidrule(lr){7-7}
\cmidrule(lr){8-8}
 & & RAOS & ToothFairy2 & MRISegmenter
   & MoCA-Mask & LVIS-92i & iSAID-5i \\
\midrule
\multicolumn{2}{@{}l}{Default}
 & 0.553 & 0.016 & 0.463 & 0.351 & 0.545 & 0.389 \\
\midrule
\multirow{3}{*}{Prompt learning}
 & Geometry embedding
 & 0.627\up & 0.163\up & 0.462 & 0.351 & 0.548 & 0.403\up \\
 & FS-CPL~\cite{venkataramani2026fscpl}
 & 0.626\up & 0.113\up & 0.487\up & 0.392 & 0.441\dn & 0.278\dn \\
 & M2C~\cite{zhou2026m2c}
 & 0.570\up & 0.236\up & 0.458 & 0.563\up & 0.740\up & 0.673\up \\
\midrule
\multirow{2}{*}{Reference-based}
 & PerSAM~\cite{zhang2024persam} (original)
 & 0.425\dn & 0.109\up & 0.453 & 0.379 & 0.642\up & 0.539\up \\
 & PerSAM~\cite{zhang2024persam} (enhanced)
 & 0.548\dn & 0.101\up & 0.493\up & 0.395 & 0.637\up & 0.521\up \\
\midrule
Fine-tuning & LoRA~\cite{hu2022lora}
 & 0.593\up & \underline{0.287}\up & 0.490\up & \textbf{0.639}\up & \underline{0.755}\up & \textbf{0.721}\up \\
\midrule
\multirow{3}{*}{Re-scoring}
 & Global threshold
 & 0.520\dn & 0.031\up & 0.435\dn & 0.267\dn & 0.602\up & 0.435\up \\
 & Per-class threshold
 & 0.531\dn & 0.035\up & 0.434\dn & 0.267\dn & 0.606\up & 0.436\up \\
 & PrAda~\cite{rosi2026prada}
 & 0.641\up & 0.062\up & \underline{0.534}\up & 0.634\up & 0.388\dn & 0.219\dn \\
\midrule
\multirow{2}{*}{Steering}
 & Intra-modal steering
 & 0.630\up & 0.131\up & 0.434\dn & 0.557\up & 0.580\up & 0.488\up \\
 & \textbf{\ours (ours)}
 & \underline{0.642}\up & 0.132\up & 0.513\up & 0.622\up & 0.657\up & 0.523\up \\
\midrule
\multirow{2}{*}{\textit{Hybrid}}
 & LoRA + \textbf{\ours (ours)}
 & 0.634\up & \textbf{0.291}\up & 0.507\up & \underline{0.638}\up & \textbf{0.759}\up & \underline{0.718}\up \\
 & PrAda + \textbf{\ours (ours)}
 & \textbf{0.658}\up & 0.117\up & \textbf{0.549}\up & 0.625\up & 0.454\dn & 0.222\dn \\
\bottomrule
\end{tabular}}
\caption{Comparison with few-shot adaptation baselines
and their combinations with \ours, using five annotated images per class.} 
\label{tab:main}
\end{table*}

\subsection{Understanding Cross-Modal Steering}

\paran{Cross-modal vs. intra-modal steering}
To isolate the benefit of cross-modal guidance,
we replace the optimised click with an optimised
text embedding, retaining the same steering protocol.
This adapts \textit{intra-modal steering} to segmentation.
As shown in \Cref{tab:main}, cross-modal steering
shows stronger performance on all six datasets,
with particularly large gains on MRISegmenter
(+7.9), LVIS-92i (+7.7),
and MoCA-Mask (+6.5).
These results highlight the importance of
cross-modal guidance beyond activation steering alone.

\begin{figure}[t]
\centering
\includegraphics[width=\columnwidth, trim=0 10pt 0 5pt, clip]{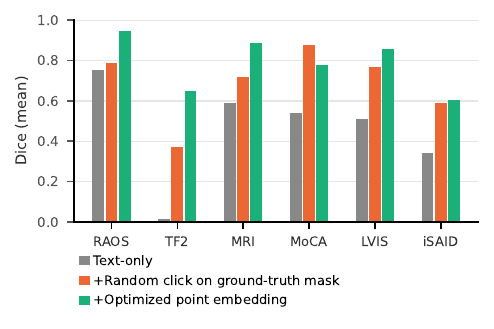}
\caption{Ablation on our prompt optimization strategy.
}
\label{fig:point-optimization}
\end{figure}

\para{Prompt optimization}
We further ablate our prompt optimising step, to verify that it enables stronger cross-modal guidance than clicks sampled from the ground-truth mask.
As shown in \Cref{fig:point-optimization},
optimisation improves segmentation
across most datasets (except for MoCA).
Gains are particularly large
on ToothFairy2 and MRISegmenter.
We can also see the positive impact
over raw clicks on RAOS (0.642 vs. 0.552), ToothFairy2 (0.132 vs. 0.063 and MRISegmenter (0.513 vs 0.486) when comparing results with one click in \Cref{tab:no-masks} against \ours in \Cref{tab:main}.
Overall, this shows that the effectiveness of cross-modal steering depends on the guidance from which its vectors are derived.

\para{Auto-Steering configuration}
\Cref{tab:auto-steering} investigates how steering
performance varies with the intervention site and strength.
While configurations sometimes transfer between related
datasets, such as RAOS and MRISegmenter,
others substantially degrade performance.
For example, transferring the iSAID-5i configuration
to LVIS-92i reduces Dice from 0.545 to 0.282.
Our Auto-Steering procedure instead selects suitable
configurations using only the available adaptation examples,
achieving improvements over text-only prompting
across all six datasets.

\subsection{Comparison and Combination with Baselines}
\label{sec:baseline-comparisons}

\paran{Performance under matched adaptation budget}
We compare \ours against representative few-shot adaptation
methods using the same five annotated images per class
(\cref{tab:main}).
\ours achieves competitive performance across the six datasets,
outperforming most baselines on RAOS and MRISegmenter.
LoRA and M2C achieve higher Dice scores on non-medical datasets, 
which does not require the model to refrain from predictions.
This is required in volumes in the medical domain, which often have the target class only on some slices
(further analysis on absent classes below).
Unlike parameter-based fine-tuning, \textit{\ours adapts
internal activations without updating model weights},
requiring substantially fewer floating-point operations (FLops)
than LoRA (\Cref{sup:cost} in the supplement).

\para{Complementarity with existing approaches}
We further investigate whether cross-modal steering
can complement existing adaptation methods.
Combining LoRA with \ours improves performance on
four of six datasets.
On the other two datasets, cross-modal guidance
itself is not improving segmentation for LoRA
(\Cref{sup:clickopt}).
Combining PrAda with \ours yields further gains
on RAOS (0.641 to 0.658) and MRISegmenter
(0.534 to 0.549),
achieving the overall best performance on these datasets.

\begin{table}[t]
\centering
\small
\setlength{\tabcolsep}{3pt}
\resizebox{\columnwidth}{!}{%
\begin{tabular}{@{}lcccccc@{}}
\toprule
 & \multicolumn{3}{c}{LVIS-92i}
 & \multicolumn{3}{c}{iSAID-5i} \\
\cmidrule(lr){2-4}\cmidrule(lr){5-7}
 & Present & Present (Top-1) & Absent
 & Present & Present (Top-1) & Absent \\
\midrule
M2C
 & 81.3 & 81.8 & 54.9
 & 69.9 & 74.4 & 48.0 \\
LoRA
 & \textbf{83.9} & \textbf{84.0} & 36.6
 & \textbf{80.3} & \textbf{80.3} & 16.3 \\
\textbf{\ours (ours)}
 & 79.2 & 83.7 & \textbf{84.5}
 & 46.4 & 64.3 & \textbf{78.6} \\
\bottomrule
\end{tabular}}
\caption{Segmentation of present and rejection of absent
classes (in \%). Top-1: Dice when retaining the highest-scoring
mask even below the confidence threshold.
Absent: correct empty predictions for challenging
negative prompts.}
\label{tab:absent-main}
\end{table}

\para{Recognizing absent classes}
Few-shot adaptation improve segmentation of present
classes at the cost of predicting absent structures (\ie false positives).
\Cref{tab:absent-main} illustrates this trade-off on negative prompts
(details in \Cref{sup:absent} in the supplement).
While LoRA and M2C achieve higher Dice for present classes,
their correctness when the class is absent drops substantially,
reaching only 16.3\% for LoRA on iSAID-5i.
In contrast, \ours retains 84.5\% and 78.6\%
correctness
on LVIS-92i and iSAID-5i, respectively.
The results for the top-1 mask (without threshold-based pruning)
show that many of the remaining errors of \ours stem from thresholding,
preventing false positives.

%% file: sec/6_conclusion.tex
\section{Conclusion}
\label{sec:conclusion}

Cross-modal steering offers a different perspective on
adapting multimodal models:
rather than updating model parameters,
we can exploit capabilities elicited by additional
modalities and transfer their effects to inference
without those modalities.
Our results show that this principle works across
different segmentation architectures, prompt modalities, supervision settings,
and even enables test-time adaptation.
By enabling adaptation from only a few annotations
or even sparse clicks, cross-modal steering offers
a practical way to specialize existing models
without costly fine-tuning or additional inference-time guidance.

Code and demos are publicly available.
Due to simple setup, consistent performance gains, and low computational cost, we expect wide-spread interest and adoption of our method among researchers and practitioners.

%% file: sec/X_suppl.tex
\clearpage
\setcounter{section}{0}
\renewcommand{\thesection}{S\arabic{section}}
\setcounter{table}{0}\renewcommand{\thetable}{S\arabic{table}}
\setcounter{figure}{0}\renewcommand{\thefigure}{S\arabic{figure}}
\maketitlesupplementary
\raggedbottom
\setlength{\intextsep}{8pt plus 2pt minus 2pt}

\input{sec/supplementary/1_limitations}

\input{sec/supplementary/2_expt_setup}
\input{sec/supplementary/3_further_analysis}

\input{sec/supplementary/4_indepth_results}
\input{sec/supplementary/5_anecdotes}

\input{sec/supplementary/6_cost}

%% file: sec/supplementary/1_limitations.tex
\section{Limitations and Future Work}
\label{sup:lim_future}

\paran{Site selection}
Auto-Steering selects the intervention site with the
largest mean leave-one-out Dice gain on the available
adaptation examples.
However, with limited supervision, this estimate can
be sensitive to individual examples with large gains.
Moreover, the validation gain provides a ranking signal,
but does not reliably predict test performance
(\cref{tab:sites}).
For instance, with five ground-truth clicks per support image instead of segmentation masks, \ours achieves only 0.435 Dice
on MRISegmenter, falling below text-only prompting (0.463).
Using the site selected with segmentation masks instead
improves Dice to 0.480, suggesting that site selection
accounts for much of this difference.
More robust selection criteria, considering the variance
or consistency of gains across examples, could help
address this limitation.
We also leave steering at multiple sites, jointly or
sequentially, for future work.

\para{Limits of cross-modal guidance}
Cross-modal steering relies on the additional modality
providing useful information beyond the inference prompt.
However, additional guidance does not always improve
segmentation.
On MoCA-Mask, providing an additional text prompt does not improve performance over click-only inference (0.659 vs.\ 0.634, \cref{tab:clicktransfer_nat}).
Consequently, steering click-only inference with
text-derived vectors reduces Dice to 0.592.
These results highlight that the effectiveness of
cross-modal steering depends on the benefit of
the additional modality.

\para{Suppressing false positives}
Our steering vectors are derived from positive clicks
and primarily aim to recover missed structures.
While \ours preserves absent-class recognition
better than several adaptation baselines
(\cref{tab:absent_new}), it does not explicitly
target false-positive predictions.
Steering vectors derived from negative clicks,
or combining positive and negative guidance,
could help address this limitation.

\para{Prompt optimization objective}
Our prompt optimization minimizes the soft Dice loss
on masks above the confidence threshold, or on the
highest-scoring mask if none passes the threshold.
However, confidence scores are not optimized,
so improved masks may still be discarded at inference.
Incorporating confidence or presence prediction
into the optimization objective could address this limitation.

\para{Scope and generality}
While our experiments focus on promptable segmentation,
the principle of cross-modal steering is not restricted
to this setting.
Future work could explore its applicability to other
multimodal models and tasks, transferring information
from additional modalities without requiring them at inference.

%% file: sec/supplementary/2_expt_setup.tex
\section{Experimental Setup}

\subsection{Details on Datasets}
\label{sup:datasets}

\Cref{tab:datasets} summarizes the six datasets.
All methods use the same five annotated examples per class
and identical class-specific text prompts.
These prompts are used for text-only inference,
steering vector derivation, and baseline adaptation.

\begin{table*}[t]
\centering
\small
\setlength{\tabcolsep}{3pt}
\resizebox{\linewidth}{!}{%
\begin{tabular}{@{}lllllll@{}}
\toprule
 & Model & Classes & Text prompt & Support (five per class) & Test & Metric \\
\midrule
RAOS & Medical-SAM3 & 19 organs & organ name (\cref{tab:prompts_raos}) & middle slice of each class in 5 training volumes & 67 test volumes & Dice, NSD \\
ToothFairy2 & Medical-SAM3 & 14 label groups & group name (\cref{tab:tf2groups}) & middle slice of each class in 5 volumes & 48 volumes & Dice, NSD \\
MRISegmenter & Medical-SAM3 & 33 structures & structure name (\cref{tab:prompts_mriseg}) & middle slice of each class in 5 training volumes & 60 test volumes & Dice, NSD \\
MoCA-Mask & SAM3 & 1 (animal) & ``animal'' & middle annotated frame of 5 training videos & 16 videos, 745 frames & Dice per video \\
LVIS-92i & SAM3 & 276 (folds 0--2) & LVIS category name & 5 LVIS-train images & 6{,}900 episodes & Dice per episode \\
iSAID-5i & SAM3 & 15 (3 folds of 5) & iSAID class name & 5 iSAID-train patches & 10{,}063 (patch, class) pairs & Dice per pair \\
\bottomrule
\end{tabular}}
\caption{Datasets. Support sets are disjoint from the test sets.}
\label{tab:datasets}
\end{table*}

\para{Medical datasets}
We process medical volumes slice by slice in the axial plane
and report class-mean Dice and NSD per volume.
RAOS~\cite{luo2024raos} CT volumes are windowed to
$[-160,240]$\,HU, while MRISegmenter~\cite{zhuang2025mrisegmenter}
volumes are clipped to their
$[0.5,99.5]$ intensity percentiles.

Each class is prompted using its name alone,
following the first phrasing in the dataset's prompt list for consistency
(\eg ``Liver'' rather than ``Liver in abdominal CT'').
For adaptation, we use one annotated 2D mask per class
and support volume, corresponding to the middle axial
slice containing the target structure.
No additional annotations, including class-absence labels,
are provided.

For RAOS, we use five training volumes from Set~A
(patients without organ resection) for adaptation
and evaluate on all 67 test volumes from the same set.
Sets~B and C, containing patients with partial or
complete organ resection, are excluded.
For MRISegmenter, we use support volumes
\texttt{train\_017, 075, 085, 097, 105}
and evaluate on the 60 venous test volumes.
For ToothFairy2~\cite{bolelli2026toothfairy2},
we use support volumes
\texttt{P\_047, P\_084, P\_369, P\_452, F\_045}
and 48 separate test volumes.

\para{ToothFairy2 label groups}
ToothFairy2 annotates 42 classes, including individual
teeth identified by FDI notation.
We merge these into 14 groups (\cref{tab:tf2groups})
to reduce adaptation costs across methods
and avoid left/right ambiguities in text prompts.
Teeth are grouped by jaw and type, while the
supernumerary tooth (label 40) is treated as background.
Ground-truth and predicted masks are evaluated
using these 14 groups.

\para{MoCA-Mask}
MoCA-Mask is a binary video segmentation benchmark featuring camouflaged animals.
We therefore use the same text prompt, ``animal'',
for all videos.
For adaptation, we select the middle annotated frame
from five training videos, following the official
training split (seed 42).
We evaluate on all 16 test videos, comprising
745 annotated frames.
Dice is computed per frame and averaged first
over frames within each video, then across videos.

\para{LVIS-92i}
We follow the LVIS-92i protocol of Matcher~\cite{liu2024matcher},
which divides 920 classes into ten folds of 92 classes each,
using LVIS validation images as queries
(2{,}300 episodes per fold).
We evaluate folds 0--2, comprising 276 classes and
6{,}900 episodes.
Text prompts use the LVIS category names,
replacing underscores with spaces while retaining
disambiguations (\eg ``bow (weapon)'').
For adaptation, we randomly select five LVIS training
images per class, using the union of all instances
of that class as the support mask.
Classes with fewer than five training images
(18 of 920) use all available examples.
The steering site and strength $\alpha$ are selected
separately for each fold across its 92 classes.

\para{iSAID-5i}
iSAID-5i~\cite{zamir2019isaid} divides the 15 iSAID classes
into three folds of five classes each, using
$256\times256$ image patches.
We follow protocol A of GF-SAM~\cite{zhang2024gfsam},
where each (validation patch, present class) pair
forms a query.
This yields 3{,}108, 3{,}657, and 3{,}298 queries
for folds 0--2, respectively, evaluated at 1024\,px.
Text prompts use the lower-case class names:
ship, storage tank, baseball diamond, tennis court,
basketball court, ground track field, bridge,
large vehicle, small vehicle, helicopter, swimming pool,
roundabout, soccer ball field, plane, and harbor.
For adaptation, we select five training patches per class.
The steering site and strength $\alpha$ are selected
separately for each fold across its five classes.

\begin{table}[t]
\centering
\small
\setlength{\tabcolsep}{4pt}
\resizebox{\linewidth}{!}{%
\begin{tabular}{@{}llll@{}}
\toprule
Group (prompt) & Original labels & Group (prompt) & Original labels \\
\midrule
Lower Jawbone & 1 & Upper Incisor & 11, 12, 21, 22 \\
Upper Jawbone & 2 & Upper Canine & 13, 23 \\
Inferior Alveolar Canal & 3, 4 (left, right) & Upper Premolar & 14, 15, 24, 25 \\
Maxillary Sinus & 5, 6 (left, right) & Upper Molar & 16, 17, 18, 26, 27, 28 \\
Pharynx & 7 & Lower Incisor & 31, 32, 41, 42 \\
Crown, Bridge or Implant & 8, 9, 10 & Lower Canine & 33, 43 \\
 & & Lower Premolar & 34, 35, 44, 45 \\
 & & Lower Molar & 36, 37, 38, 46, 47, 48 \\
\bottomrule
\end{tabular}}
\caption{ToothFairy2 groups and their prompts (FDI tooth notation).}
\label{tab:tf2groups}
\end{table}

\begin{table}[t]
\centering
\small
\begin{tabular}{@{}p{0.97\linewidth}@{}}
\toprule
Liver, Spleen, Left kidney, Right kidney, Stomach, Gallbladder, Esophagus, Pancreas, Duodenum, Colon, Intestine, Right adrenal gland, Left adrenal gland, Rectum, Bladder, Left head of femur, Right head of femur, Prostate, Seminal vesicles \\
\bottomrule
\end{tabular}
\caption{Text prompts of RAOS (19 classes).}
\label{tab:prompts_raos}
\end{table}

\begin{table}[t]
\centering
\small
\begin{tabular}{@{}p{0.97\linewidth}@{}}
\toprule
Spleen, Right kidney, Left kidney, Gallbladder, Liver, Esophagus, Stomach, Aorta, Inferior vena cava, Portal vein and splenic vein, Pancreas, Right adrenal gland, Left adrenal gland, Right lung, Left lung, Small bowel, Duodenum, Colon, Left iliac artery, Right iliac artery, Left iliac vein, Right iliac vein, Left gluteus maximus, Right gluteus maximus, Left gluteus medius, Right gluteus medius, Left autochthon, Right autochthon, Left iliopsoas, Right iliopsoas, Left hip, Right hip, Sacrum \\
\bottomrule
\end{tabular}
\caption{Text prompts of MRISegmenter (33 classes).}
\label{tab:prompts_mriseg}
\end{table}

\subsection{Selected Sites and Strengths}

Auto-Steering selects intervention sites and strengths
using only the available adaptation examples.
We investigate how well the resulting leave-one-out
validation gains reflect performance on the test sets.

\Cref{tab:sites} reports the selected sites, steering
strengths $\alpha$, and Dice before and after steering
on the held-out adaptation examples.
For LVIS-92i and iSAID-5i, configurations are selected
per fold, pooling over its classes.
The validation gain serves as a ranking signal,
but does not reliably predict the test gain.
It overestimates improvements on most datasets,
including ToothFairy2 (+0.203 vs.\ +0.116)
and MRISegmenter (+0.092 vs.\ +0.049),
while underestimating them on RAOS
(+0.055 vs.\ +0.089) and iSAID-5i fold~2.

\begin{table}[t]
\centering
\small
\setlength{\tabcolsep}{5pt}
\resizebox{\linewidth}{!}{%
\begin{tabular}{@{}lllcc@{}}
\toprule
 & Site $\ell$ & $\alpha$ & Annotated examples & Test $\Delta$Dice \\
\midrule
RAOS & encoder.0.cross\_attn & 1.68 & 0.751$\rightarrow$0.806 (5/5) & +0.089 \\
ToothFairy2 & encoder.1 & 2.79 & 0.015$\rightarrow$0.218 (5/5) & +0.116 \\
MRISegmenter & encoder.0.cross\_attn & 1.84 & 0.591$\rightarrow$0.683 (5/5) & +0.049 \\
\midrule
MoCA-Mask & encoder.4 & 3.24 & 0.538$\rightarrow$0.871 (2/5) & +0.271 \\
LVIS-92i, fold 0 & encoder.3.self\_attn & 3.15 & 0.531$\rightarrow$0.667 (237/452) & +0.102 \\
LVIS-92i, fold 1 & encoder.3.self\_attn & 3.11 & 0.544$\rightarrow$0.654 (206/459) & +0.112 \\
LVIS-92i, fold 2 & encoder.3.self\_attn & 3.21 & 0.549$\rightarrow$0.672 (211/453) & +0.121 \\
iSAID-5i, fold 0 & encoder.1 & 3.61 & 0.332$\rightarrow$0.553 (11/25) & +0.198 \\
iSAID-5i, fold 1 & encoder.3 & 3.68 & 0.326$\rightarrow$0.482 (9/25) & +0.048 \\
iSAID-5i, fold 2 & encoder.3.self\_attn & 2.64 & 0.362$\rightarrow$0.446 (8/25) & +0.169 \\
\bottomrule
\end{tabular}}
\caption{Selected site and strength $\alpha$ of \ours, leave-one-out Dice on the annotated examples (improved / all held-out units) and test gain over the text prompt.}
\label{tab:sites}
\end{table}

%% file: sec/supplementary/3_further_analysis.tex
\section{Further Analysis}

\subsection{Adaptation Budget}
\begin{figure}[t]
\centering
\includegraphics[width=\columnwidth, trim=0 9pt 0 10pt, clip]{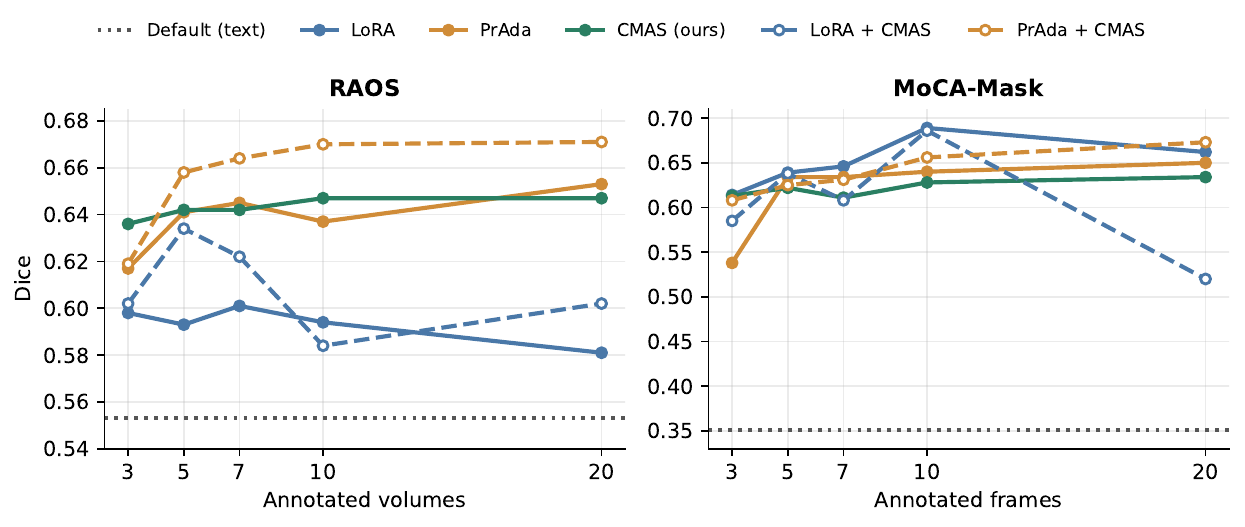}
\caption{Test Dice for different adaptation budgets; dotted line: text-only.}
\label{fig:budget}
\end{figure}

We investigate how the number of annotated examples
affects adaptation performance.
\Cref{fig:budget} compares \ours against LoRA and PrAda,
as well as their combinations, on RAOS and MoCA-Mask.
With only three annotated examples, \ours achieves
the best performance among the individual methods
on both datasets.
Moreover, its performance improves consistently
with additional annotations, while other approaches
exhibit larger fluctuations or even performance drops.
These results suggest that cross-modal steering
provides a robust adaptation strategy,
particularly under very limited annotation budgets.

\subsection{Error Analysis}
\label{sup:errors}

To better understand how different adaptation methods
improve segmentation, we examine which failure modes
they correct and which errors remain.
\Cref{tab:errors_cmas_lora,tab:errors_cmas_lora_nat}
decompose the Dice deficit of \ours, LoRA, and M2C
into the failure modes introduced in \Cref{sec:diagnosis}.

On the natural-image datasets, \ours primarily reduces
missed objects, with presence FN decreasing from
0.573 to 0.089 on MoCA-Mask, from 0.352 to 0.165
on LVIS-92i, and from 0.447 to 0.242 on iSAID-5i.
On RAOS, the largest improvement comes from reducing
semantic mismatches (0.102 to 0.045), particularly
confusion between left and right structures.

However, correcting one failure mode can introduce
others.
On ToothFairy2, \ours reduces missed structures,
but semantic mismatches increase from 0.036 to 0.157,
suggesting that some previously missed structures
are now incorrectly identified.
LoRA and M2C exhibit similar behavior, with semantic
mismatches increasing to 0.178 and 0.187, respectively.

More generally, LoRA and M2C substantially reduce
presence false negatives, but often increase
false-positive presence predictions.
For example, on MRISegmenter, presence FP increases
from 0.248 to 0.366 with LoRA and 0.340 with M2C,
compared to 0.259 with \ours.
These results highlight differences in how adaptation
methods trade missed structures against false-positive
predictions.

\begin{table}[t]
\centering
\small
\setlength{\tabcolsep}{3pt}
\resizebox{\linewidth}{!}{%
\begin{tabular}{@{}l|cccc|cccc|cccc@{}}
\toprule
 & \multicolumn{4}{c}{RAOS} & \multicolumn{4}{c}{ToothFairy2} & \multicolumn{4}{c}{MRISeg.} \\
\cmidrule(lr){2-5}\cmidrule(lr){6-9}\cmidrule(lr){10-13}
 & Text & \ours & LoRA & M2C & Text & \ours & LoRA & M2C & Text & \ours & LoRA & M2C \\
\midrule
Dice & 0.553 & 0.642 & 0.593 & 0.570 & 0.016 & 0.132 & 0.287 & 0.236 & 0.464 & 0.513 & 0.490 & 0.458 \\
\midrule
Presence FN & 0.045 & 0.031 & 0.014 & 0.018 & 0.419 & 0.269 & 0.035 & 0.047 & 0.085 & 0.055 & 0.021 & 0.031 \\
Presence FP & 0.180 & 0.181 & 0.227 & 0.272 & 0.404 & 0.353 & 0.457 & 0.459 & 0.248 & 0.259 & 0.366 & 0.340 \\
Semantic mismatch & 0.102 & 0.045 & 0.056 & 0.044 & 0.036 & 0.157 & 0.178 & 0.187 & 0.109 & 0.076 & 0.045 & 0.084 \\
FN area & 0.105 & 0.085 & 0.089 & 0.080 & 0.080 & 0.055 & 0.027 & 0.040 & 0.064 & 0.064 & 0.055 & 0.063 \\
FP area & 0.013 & 0.015 & 0.020 & 0.016 & 0.045 & 0.033 & 0.016 & 0.030 & 0.029 & 0.032 & 0.023 & 0.023 \\
\bottomrule
\end{tabular}}
\caption{Dice deficit decomposed into the five failure modes
for text-only prompting and three adaptation methods
on medical datasets. Lower is better for error contributions.}
\label{tab:errors_cmas_lora}
\end{table}

\begin{table}[t]
\centering
\small
\setlength{\tabcolsep}{3pt}
\resizebox{\linewidth}{!}{%
\begin{tabular}{@{}l|cccc|cccc|cccc@{}}
\toprule
 & \multicolumn{4}{c}{MoCA-Mask} & \multicolumn{4}{c}{LVIS-92i} & \multicolumn{4}{c}{iSAID-5i} \\
\cmidrule(lr){2-5}\cmidrule(lr){6-9}\cmidrule(lr){10-13}
 & Text & \ours & LoRA & M2C & Text & \ours & LoRA & M2C & Text & \ours & LoRA & M2C \\
\midrule
Dice & 0.351 & 0.622 & 0.639 & 0.563 & 0.545 & 0.657 & 0.755 & 0.740 & 0.389 & 0.523 & 0.721 & 0.673 \\
\midrule
Presence FN & 0.573 & 0.089 & 0.100 & 0.274 & 0.352 & 0.165 & 0.029 & 0.065 & 0.447 & 0.242 & 0.009 & 0.060 \\
Semantic mismatch & NA & NA & NA & NA & 0.019 & 0.044 & 0.045 & 0.037 & 0.040 & 0.071 & 0.068 & 0.062 \\
FN area & 0.022 & 0.143 & 0.088 & 0.060 & 0.049 & 0.061 & 0.081 & 0.085 & 0.066 & 0.084 & 0.148 & 0.132 \\
FP area & 0.053 & 0.147 & 0.174 & 0.103 & 0.035 & 0.073 & 0.090 & 0.073 & 0.058 & 0.080 & 0.054 & 0.073 \\
\bottomrule
\end{tabular}}
\caption{As \cref{tab:errors_cmas_lora}, natural-image datasets (no presence FP: the class is present in every frame or query). NA: not applicable (MoCA-Mask has one class).}
\label{tab:errors_cmas_lora_nat}
\end{table}

\subsection{What Prompt Optimization Changes}
\label{sup:clickopt}

\begin{figure}[t]
\centering
\includegraphics[width=\linewidth]{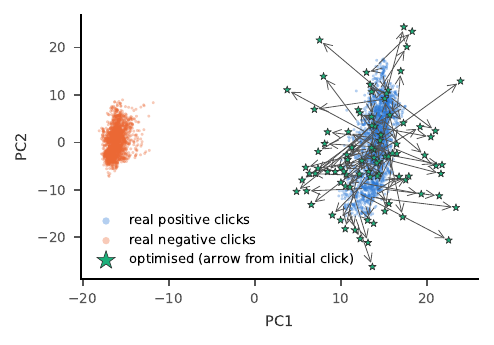}
\caption{Click prompt tokens of real and optimised clicks on RAOS (PCA).}
\label{fig:clickopt}
\end{figure}

\ours optimizes point-prompt embeddings before deriving
cross-modal steering vectors (\cref{sec:method}).
While \Cref{fig:point-optimization} shows the benefits
for segmentation performance, we investigate how
the optimized embeddings differ from those of real clicks.
We analyze the middle slice of each class in the first
five test volumes of RAOS and ToothFairy2
(89 and 66 slices, respectively).
These examples are used for analysis only,
without influencing method selection or configuration.
For each slice, we extract embeddings of 20 positive
clicks inside the ground-truth mask and 20 negative clicks
in a ring 3--8 pixels outside it.
We then project these embeddings, together with the
optimized point embeddings, into two dimensions using PCA
(\cref{fig:clickopt}).

\para{Optimized embeddings differ from real clicks}
To understand how optimization changes the prompt,
we compare the optimized embeddings against those
of real positive and negative clicks.
The latter are perfectly linearly separable on both
datasets (linear-probe accuracy 1.00), with approximately
98\% of their separation explained by the click label
rather than its location.
Optimized embeddings remain on the positive side,
but move substantially beyond the space occupied
by real clicks (\cref{fig:clickopt}).
Their displacement perpendicular to the positive--negative
axis is 20--34 units, compared to only 1.5--1.7 units
along it.
Moreover, their distance to the nearest real positive
click is 5.6--6.4 times the typical spacing between
real clicks, while the geometry CLS token moves
even further (32--38 times).
At the steering site, the resulting activation
differences retain the positive-click classification
on 93--97\% of slices.

\para{Implications for cross-modal steering}
The optimized embeddings therefore do not necessarily
correspond to clicks that a user could provide.
Instead, optimization explores the continuous prompt
embedding space to obtain stronger corrective guidance
for deriving steering vectors.
As shown in the main paper, this generally improves
steering performance on the medical datasets,
although raw-click vectors perform slightly better
in MoCA-Mask and on par in iSAID-5i
(\cref{fig:point-optimization}).

\begin{figure}[t]
\centering
\includegraphics[width=\linewidth, trim=0 10pt 0 5pt, clip]{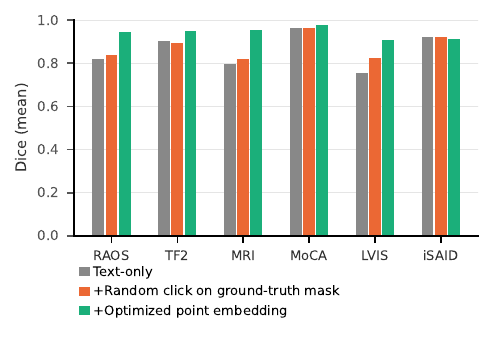}
\caption{As \cref{fig:point-optimization}, on top of the same-budget LoRA models (Dice on the annotated support examples).}
\label{fig:clickopt_lora}
\end{figure}

\para{Prompt optimization after fine-tuning}
We further investigate whether prompt optimization
remains beneficial after parameter-efficient fine-tuning,
when the model already fits the annotated examples well.
\Cref{fig:clickopt_lora} shows the results for this analysis on LoRA models.
The trained LoRA models already achieve high Dice on their support examples (0.75--0.97).
As a result, a random click provides little additional
benefit ($-0.01$ to $+0.07$).
However, optimizing the point embeddings still yields
substantial improvements on four of six datasets,
including RAOS (0.82 to 0.95), MRISegmenter
(0.80 to 0.95), and LVIS-92i (0.75 to 0.91).
On MoCA-Mask, performance remains similar, while
iSAID-5i shows a slight decrease (0.923 to 0.910),
partly due to optimized predictions falling below
the confidence threshold.
These results show that prompt optimization can
provide additional corrective guidance even after
fine-tuning.
However, effects are smaller than observed in \Cref{fig:clickopt}.

\subsection{Robustness to Support Selection}

With only five annotated examples per class,
adaptation performance may depend on which examples
are selected.
We therefore repeat the full adaptation procedure,
including site selection, with two additional random
draws of support volumes, videos, or images.
As shown in \Cref{tab:draws}, \ours improves over
text-only prompting in all 18 runs, significantly
in 17 of them.
Improvements remain relatively stable on the medical
datasets, LVIS-92i, and iSAID-5i, despite differences
in the selected steering sites.
MoCA-Mask exhibits the largest variation
(+0.139 to +0.271 Dice).

\begin{table}[t]
\centering
\small
\setlength{\tabcolsep}{4pt}
\begin{tabular}{@{}lccc@{}}
\toprule
$\Delta$Dice vs. Text & Draw 0 (main) & Draw 1 & Draw 2 \\
\midrule
RAOS & +0.089\up & +0.085\up & +0.084\up \\
ToothFairy2 & +0.116\up & +0.075\up & +0.109\up \\
MRISegmenter & +0.049\up & +0.031\up & +0.048\up \\
\midrule
MoCA-Mask & +0.271\up & +0.139 & +0.236\up \\
LVIS-92i & +0.112\up & +0.103\up & +0.117\up \\
iSAID-5i & +0.134\up & +0.100\up & +0.095\up \\
\bottomrule
\end{tabular}
\caption{\ours with three random draws of the five annotated volumes or images per class. \up: better than the text prompt (paired Wilcoxon, $p<0.05$).}
\label{tab:draws}
\end{table}

%% file: sec/supplementary/4_indepth_results.tex
\section{In-depth Results}

\subsection{Normalized Surface Dice}
\label{sup:nsd}

In addition to Dice, we evaluate boundary accuracy
using Normalized Surface Dice (NSD).
NSD measures surface agreement within a tolerance
defined in millimetres, accounting for voxel spacing.
We report NSD only for medical datasets, as natural-image
benchmarks do not provide physical pixel spacing.

\Cref{tab:nsd} shows that \ours improves NSD over
text-only prompting on all three medical datasets:
RAOS (0.561 to 0.640), ToothFairy2 (0.059 to 0.242),
and MRISegmenter (0.459 to 0.513).
The best-performing methods are consistent with
the Dice results in \Cref{tab:main}:
PrAda + \ours on RAOS and MRISegmenter,
and LoRA + \ours on ToothFairy2.
These results indicate that the improvements
also extend to segmentation boundary accuracy.

\begin{table}[t]
\centering
\small
\setlength{\tabcolsep}{5pt}
\resizebox{\linewidth}{!}{%
\begin{tabular}{@{}llccc@{}}
\toprule
 & & \multicolumn{3}{c}{\textbf{Medical} (Medical-SAM3)} \\
\cmidrule(lr){3-5}
 & & RAOS & ToothFairy2 & MRISegmenter \\
\midrule
\multicolumn{2}{@{}l}{Default} & 0.561 & 0.059 & 0.459 \\
\midrule
\multirow{3}{*}{Prompt learning}
 & Geometry embedding & 0.588 & 0.234 & 0.466 \\
 & FS-CPL~\cite{venkataramani2026fscpl} & 0.635 & 0.165 & 0.477 \\
 & M2C~\cite{zhou2026m2c} & 0.529 & 0.371 & 0.458 \\
\midrule
\multirow{2}{*}{Reference-based}
 & PerSAM~\cite{zhang2024persam} (original) & 0.306 & 0.255 & 0.409 \\
 & PerSAM~\cite{zhang2024persam} (enhanced) & 0.555 & 0.253 & 0.492 \\
\midrule
Fine-tuning & LoRA~\cite{hu2022lora} & 0.558 & \underline{0.432} & 0.486 \\
\midrule
\multirow{3}{*}{Re-scoring}
 & Global threshold & 0.491 & 0.073 & 0.416 \\
 & Per-class threshold & 0.521 & 0.091 & 0.424 \\
 & PrAda~\cite{rosi2026prada} & \underline{0.649} & 0.158 & 0.508 \\
\midrule
\multirow{2}{*}{Steering}
 & Intra-modal steering & 0.627 & 0.252 & 0.412 \\
 & \textbf{\ours (ours)} & 0.640 & 0.242 & \underline{0.513} \\
\midrule
\multirow{2}{*}{\textit{Hybrid}}
 & LoRA + \textbf{\ours (ours)} & 0.610 & \textbf{0.435} & 0.512 \\
 & PrAda + \textbf{\ours (ours)} & \textbf{0.693} & 0.240 & \textbf{0.566} \\
\bottomrule
\end{tabular}}
\caption{Test NSD on the medical datasets (same runs as \cref{tab:main}).}
\label{tab:nsd}
\end{table}

\subsection{Recognizing Absent Classes}
\label{sup:absent}

Our main evaluation (\cref{tab:main}) considers only
prompts for classes present in the image.
A method can therefore gain by answering more often.
Adaptation may improve segmentation by
increasing the number of predictions, potentially
introducing false positives when the requested class
is absent.
We therefore evaluate how well different methods
recognize absent classes (\cref{tab:absent_new}).

\para{Annotation}
We sample 150 LVIS-92i test pairs (image, class).
For each image, we retrieve up to eight alternative
classes from the 276 classes in these folds,
ensuring that all methods have adaptations
for each candidate.
Candidates are ranked by cosine similarity of their
SAM3 text embeddings to the true class.

One author then inspects the candidates in order of
similarity, selecting the first class verified to be absent
from the image.
If a candidate is present, the next most similar class
is considered.
Images for which absence cannot be established are discarded.
This yields 142 images (8 discarded), each with one
challenging negative prompt
(\eg flamingo$\rightarrow$pelican,
truck$\rightarrow$school bus).

We follow the same procedure for 100 iSAID-5i
validation pairs,
using the remaining iSAID classes as candidates.
After discarding two cases, we retain 98 absent prompts,
none of which has ground-truth pixels in the corresponding patch.

\para{Evaluation protocol}
We evaluate all methods using the adaptations
obtained for \Cref{tab:main}, without refitting.
For each absent prompt, the model uses the
class-specific adaptation parameters and text prompt
of the requested class, as in regular inference.
Predictions are considered correct only if the
output mask is empty (Dice 1); otherwise, Dice is 0.

\para{Results}
\Cref{tab:absent_new} shows that \ours improves
segmentation of present classes while largely preserving
the ability to reject absent ones.
On LVIS-92i, \ours improves present-class Dice
by 12.1 points over text-only prompting, while
correctly rejecting 84.5\% of absent prompts.
In contrast, LoRA and M2C achieve higher present-class
Dice, but reject only 36.6\% and 54.9\%
of absent prompts, respectively.

Similar results are observed on iSAID-5i,
where \ours improves present-class Dice by 15.4 points
while retaining 78.6\% absent-class correctness.
LoRA and M2C reject only 16.3\% and 48.0\%
of absent prompts, respectively, despite achieving
higher Dice on present classes.
These results suggest that cross-modal steering
offers a more favorable balance between recovering
missed objects and avoiding false-positive predictions,
than these prior adaptation approaches.

Nevertheless, \ours also reduces absent-class correctness
relative to text-only prompting, and the mean of
present- and absent-class Dice does not consistently
favor a single method.

\begin{table}[t]
\centering
\small
\setlength{\tabcolsep}{4pt}
\resizebox{\linewidth}{!}{%
\begin{tabular}{@{}llcccccc@{}}
\toprule
 & & \multicolumn{3}{c}{LVIS-92i} & \multicolumn{3}{c}{iSAID-5i} \\
\cmidrule(lr){3-5}\cmidrule(lr){6-8}
 & & Present & Absent & Mean & Present & Absent & Mean \\
\midrule
Prompting & Default (text prompt) & 67.1 & 97.2 & \underline{82.1} & 31.0 & 93.9 & 62.4 \\
\midrule
\multirow{3}{*}{Prompt learning} & Geometry embedding & 66.5 & 96.5 & 81.5 & 35.6 & 94.9 & 65.3 \\
 & FS-CPL~\cite{venkataramani2026fscpl} & 53.8 & 97.2 & 75.5 & 35.9 & 98.0 & \underline{66.9} \\
 & M2C~\cite{zhou2026m2c} & 81.3 & 54.9 & 68.1 & 69.9 & 48.0 & 58.9 \\
\midrule
\multirow{2}{*}{Reference mask} & PerSAM~\cite{zhang2024persam} (original) & 75.1 & 62.0 & 68.5 & 50.4 & 82.7 & 66.5 \\
 & PerSAM~\cite{zhang2024persam} (enhanced) & 75.0 & 71.1 & 73.1 & 47.7 & 86.7 & \textbf{67.2} \\
\midrule
Fine-tuning & LoRA~\cite{hu2022lora} & \underline{83.9} & 36.6 & 60.3 & \textbf{80.3} & 16.3 & 48.3 \\
\midrule
\multirow{3}{*}{Re-scoring} & Global threshold & 71.4 & 88.0 & 79.7 & 44.8 & 83.7 & 64.2 \\
 & Per-class threshold & 72.2 & 89.4 & 80.8 & 41.6 & 85.7 & 63.7 \\
 & PrAda~\cite{rosi2026prada} & 49.0 & 98.6 & 73.8 & 25.9 & 94.9 & 60.4 \\
\midrule
\multirow{2}{*}{Steering} & Intra-modal steering & 70.5 & 95.1 & \textbf{82.8} & 50.2 & 65.3 & 57.7 \\
 & \textbf{\ours (ours)} & 79.2 & 84.5 & 81.9 & 46.4 & 78.6 & 62.5 \\
\midrule
\multirow{2}{*}{\textit{Hybrid}} & LoRA + \textbf{\ours (ours)} & \textbf{84.0} & 38.7 & 61.4 & \underline{79.9} & 29.6 & 54.7 \\
 & PrAda + \textbf{\ours (ours)} & 57.3 & 94.4 & 75.8 & 31.1 & 90.8 & 61.0 \\
\bottomrule
\end{tabular}}
\caption{Segmentation of present and rejection of absent classes for all methods (in \%), as in \Cref{tab:absent-main}. Absent: correct empty predictions for the most similar absent class (142 LVIS-92i images, 98 iSAID-5i patches). Mean: mean of the two.}
\label{tab:absent_new}
\end{table}

\subsection{Steering into Click-Only Inference}
\label{sup:clicktransfer}

Our main experiments transfer the effect of spatial
guidance to text-only inference.
To investigate whether cross-modal steering also works
in the reverse direction, we derive steering vectors
from additional text guidance and apply them during
click-only inference, without requiring class names.

\para{Steering configuration}
We contrast activations from inputs containing
both a click and an optimized class-text embedding
against inputs containing the same click alone.
The text embedding is optimized by the same prompt optimization procedure outlined in \Cref{sec:method_click}.
As in \ours, Auto-Steering selects one site and $\alpha$ per dataset (per fold on LVIS-92i and iSAID-5i) by leave-one-out on the annotated examples, here scoring click-only inference (\cref{tab:sites_click}).
We derive one steering vector per class.

\para{Evaluation protocol}
At test time, we provide one randomly sampled
positive click inside the ground-truth mask
on each slice containing the target class.
Slices without the target receive no prompt.
These clicks provide oracle information about
target presence and location.
All three prompting configurations use identical
clicks and the same class grouping as in our
main experiments.

\para{Results}
Cross-modal steering improves click-only inference
on five of six datasets
(\cref{tab:clicktransfer,tab:clicktransfer_nat}).
The largest gains occur on ToothFairy2 (+0.109 Dice),
RAOS (+0.053), and MRISegmenter (+0.040),
with improvements on nearly all test volumes.
On RAOS and ToothFairy2, steering even outperforms
direct text-plus-click prompting in Dice,
particularly by correcting semantic confusion
between related structures.
Improvements are smaller but consistent on LVIS-92i
(+0.018) and iSAID-5i (+0.031).
Only MoCA-Mask exhibits a decrease, where adding
text guidance already reduces click-only performance.

\begin{figure*}[t]
\centering
\includegraphics[width=\textwidth]{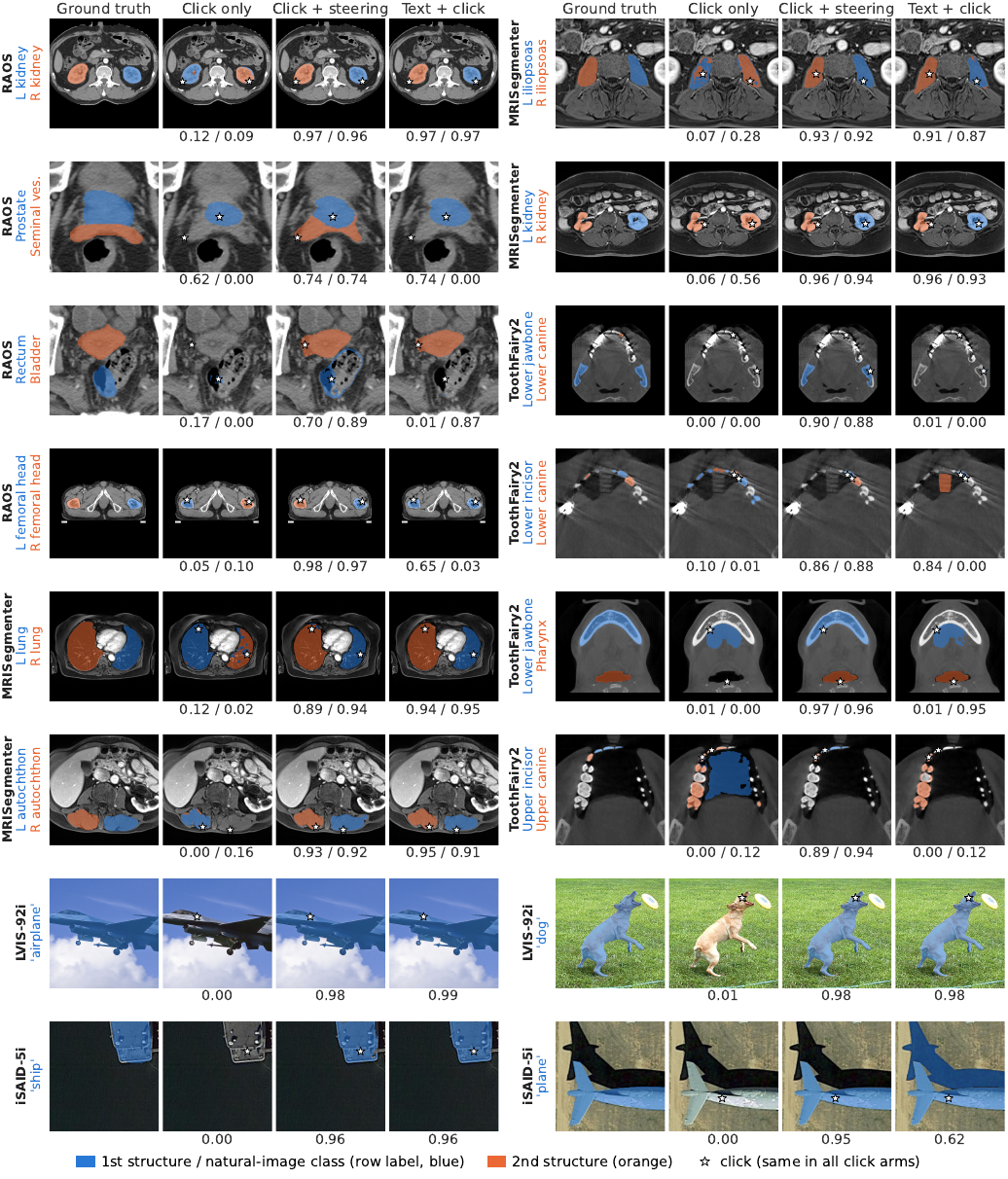}
\caption{Steering into click-only inference on RAOS, MRISegmenter, ToothFairy2, LVIS-92i and iSAID-5i; the same click (star) in all click arms. Numbers: Dice (medical: blue / orange structure).}
\label{fig:clicktransfer}
\end{figure*}

\para{Qualitative analysis}
Without a class name, a click may not sufficiently
identify the requested structure, leading to
semantic confusion or missed predictions
(\cref{fig:clicktransfer}).
For example, clicks on one kidney can produce
the opposite kidney, while clicks on the lungs
or lower jawbone sometimes yield nearly empty masks.
Text-derived steering provides additional semantic
information, allowing the same clicks to recover
the intended structures.
However, this correction is not always successful,
as illustrated by the colon example, where steering
removes an otherwise correct prediction.

\begin{table}[t]
\centering
\small
\setlength{\tabcolsep}{5pt}
\resizebox{\linewidth}{!}{%
\begin{tabular}{@{}lllcc@{}}
\toprule
 & Site $\ell$ & $\alpha$ & Annotated examples & Test $\Delta$Dice \\
\midrule
RAOS & encoder.2 & 2.48 & 0.729$\rightarrow$0.767 (4/5) & +0.053 \\
ToothFairy2 & encoder.1 & 2.38 & 0.382$\rightarrow$0.473 (5/5) & +0.109 \\
MRISegmenter & encoder.1.cross\_attn & 0.74 & 0.687$\rightarrow$0.715 (5/5) & +0.040 \\
\midrule
MoCA-Mask & encoder.1 & 2.59 & 0.886$\rightarrow$0.906 (3/5) & $-$0.067 \\
LVIS-92i, fold 0 & encoder.0.cross\_attn & 0.79 & 0.696$\rightarrow$0.719 (248/452) & +0.021 \\
LVIS-92i, fold 1 & encoder.0.cross\_attn & 0.80 & 0.690$\rightarrow$0.707 (235/459) & +0.012 \\
LVIS-92i, fold 2 & encoder.0.cross\_attn & 0.78 & 0.687$\rightarrow$0.720 (238/453) & +0.019 \\
iSAID-5i, fold 0 & encoder.3.self\_attn & 1.86 & 0.732$\rightarrow$0.752 (16/25) & +0.018 \\
iSAID-5i, fold 1 & encoder.1.mlp & 1.69 & 0.637$\rightarrow$0.652 (22/25) & +0.003 \\
iSAID-5i, fold 2 & encoder.1.self\_attn & 1.84 & 0.511$\rightarrow$0.634 (16/25) & +0.075 \\
\bottomrule
\end{tabular}}
\caption{As \cref{tab:sites}, for steering into click-only inference (\cref{tab:clicktransfer,tab:clicktransfer_nat}): leave-one-out Dice of click-only inference on the annotated examples and test gain over the click alone.}
\label{tab:sites_click}
\end{table}

\begin{table}[t]
\centering
\small
\setlength{\tabcolsep}{3pt}
\resizebox{\linewidth}{!}{%
\begin{tabular}{@{}llcccccc@{}}
\toprule
 & & \multicolumn{2}{c}{RAOS} & \multicolumn{2}{c}{ToothFairy2} & \multicolumn{2}{c}{MRISeg.} \\
\cmidrule(lr){3-4}\cmidrule(lr){5-6}\cmidrule(lr){7-8}
Prompt & Steering & Dice & NSD & Dice & NSD & Dice & NSD \\
\midrule
Text & -- & 0.553 & 0.561 & 0.016 & 0.059 & 0.463 & 0.459 \\
Text & \ours (click $\rightarrow$ text) & 0.642 & 0.640 & 0.132 & 0.242 & 0.513 & 0.513 \\
\midrule
Click & -- & 0.692 & 0.644 & 0.375 & 0.552 & 0.640 & 0.588 \\
Click & \ours (text $\rightarrow$ click) & \textbf{0.745}\up & \textbf{0.715}\up & \textbf{0.484}\up & \textbf{0.723}\up & \textbf{0.680}\up & \textbf{0.630}\up \\
\midrule
\textit{Text + click} & -- & \textit{0.733} & \textit{0.807} & \textit{0.331} & \textit{0.686} & \textit{0.729} & \textit{0.739} \\
\bottomrule
\end{tabular}}
\caption{Steering into click-only inference on the medical datasets, using five annotated images per class. \up/\dn: significantly better/worse than click only (paired Wilcoxon over volumes, videos, episodes or pairs; mIoU: over classes; $p<0.05$). Text + click is shown for reference.}
\label{tab:clicktransfer}
\end{table}

\begin{table}[t]
\centering
\small
\setlength{\tabcolsep}{3pt}
\resizebox{\linewidth}{!}{%
\begin{tabular}{@{}llcccccccc@{}}
\toprule
 & & \multicolumn{4}{c}{MoCA-Mask} & \multicolumn{2}{c}{LVIS-92i} & \multicolumn{2}{c}{iSAID-5i} \\
\cmidrule(lr){3-6}\cmidrule(lr){7-8}\cmidrule(lr){9-10}
Prompt & Steering & $S_\alpha$ & $F_\beta^w$ & $E_\phi$ & Dice & Dice & mIoU & Dice & mIoU \\
\midrule
Text & -- & 0.664 & 0.328 & 0.537 & 0.351 & 0.545 & 55.8 & 0.389 & 29.4 \\
Text & \ours (click $\rightarrow$ text) & 0.784 & 0.583 & \textbf{0.844} & 0.622 & 0.657 & \textbf{61.9} & 0.523 & 39.9 \\
\midrule
Click & -- & \textbf{0.798} & \textbf{0.607} & 0.787 & \textbf{0.659} & 0.679 & 49.0 & 0.562 & 44.0 \\
Click & \ours (text $\rightarrow$ click) & 0.713\dn & 0.544 & 0.729 & 0.592 & \textbf{0.697}\up & 52.5\up & \textbf{0.593}\up & \textbf{46.5}\up \\
\midrule
\textit{Text + click} & -- & \textit{0.798} & \textit{0.594} & \textit{0.773} & \textit{0.634} & \textit{0.773} & \textit{69.6} & \textit{0.570} & \textit{45.1} \\
\bottomrule
\end{tabular}}
\caption{As \cref{tab:clicktransfer}, natural-image datasets.}
\label{tab:clicktransfer_nat}
\end{table}

\subsection{Test-Time Adaptation}
\label{sup:tta}

We investigate whether cross-modal steering can adapt
to individual test cases without prior adaptation data.
For example, an expert may provide a few clicks on
one slice of a new patient volume, which can then
guide segmentation of the remaining slices.

\para{Protocol}
For each test volume or video, we sample three positive
clicks from the ground-truth mask on the middle slice
or frame of each present class.
We derive a case-specific steering vector from the
activation difference between text-plus-click and
text-only prompting, without prompt optimization.
The steering site and strength are taken from the
standard \ours configuration (\cref{tab:main}).
The vector is applied to the remaining slices or frames,
while the clicked image retains its directly guided
prediction.
Classes absent from a case receive no clicks
and remain text-only.

\para{Results}
\Cref{tab:tta} compares this setting against text-only
prompting and an oracle receiving three clicks
on every slice or frame.
Test-time steering improves Dice across all four datasets,
with particularly large gains on ToothFairy2
(0.016 to 0.201) and MoCA-Mask (0.351 to 0.573).
Compared to \ours using five annotated support images,
per-case steering performs better on ToothFairy2,
similarly on MRISegmenter, but worse on RAOS and MoCA-Mask.
This demonstrates the potential of cross-modal steering
for adapting individual cases with minimal interaction.

\begin{table}[t]
\centering
\small
\setlength{\tabcolsep}{4pt}
\resizebox{\linewidth}{!}{%
\begin{tabular}{@{}lccccccc@{}}
\toprule
 & \multicolumn{2}{c}{RAOS} & \multicolumn{2}{c}{ToothFairy2} & \multicolumn{2}{c}{MRISegmenter} & MoCA-Mask \\
\cmidrule(lr){2-3}\cmidrule(lr){4-5}\cmidrule(lr){6-7}\cmidrule(lr){8-8}
 & Dice & NSD & Dice & NSD & Dice & NSD & Dice \\
\midrule
Text-only prompt & 0.553 & 0.561 & 0.016 & 0.059 & 0.463 & 0.459 & 0.351 \\
Test-time adaptation with CMAS & \textbf{0.587}\up & \textbf{0.605}\up & \textbf{0.201}\up & \textbf{0.342}\up & \textbf{0.517}\up & \textbf{0.525}\up & \textbf{0.573}\up \\
\midrule
\textit{Oracle: text + 3 clicks on every slice}$^{\dagger}$ & \textit{0.649} & \textit{0.640} & \textit{0.267} & \textit{0.548} & \textit{0.597} & \textit{0.617} & \textit{0.735} \\
\bottomrule
\end{tabular}}
\caption{Test-time adaptation with per-case steering vectors from clicks on one slice or frame. \up/\dn: significantly better/worse than the text prompt (paired Wilcoxon over volumes or videos, $p<0.05$); $^{\dagger}$draw 0 of three click draws.}
\label{tab:tta}
\end{table}

\subsection{Per-Organ Gains on RAOS}

\begin{figure}[t]
\centering
\includegraphics[width=\linewidth]{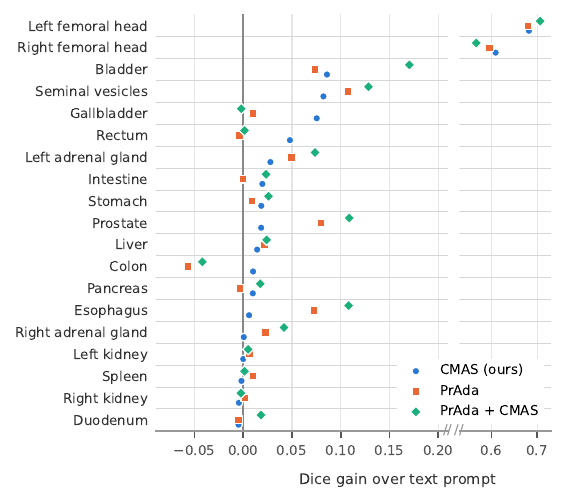}
\caption{ Per-organ test Dice gain over the text prompt on RAOS.}
\label{fig:perclass}
\end{figure}

To better understand the complementarity between
\ours and PrAda, we examine their performance
across individual organs on RAOS.
\Cref{fig:perclass} shows that the two methods
improve different structures.
While \ours provides larger gains on the gallbladder
and rectum, PrAda performs better on the prostate
and esophagus.
Their combination retains the substantial improvements
on the femoral heads and further improves the bladder,
although some gains on the gallbladder and rectum are lost.
The colon is the only structure for which the combined
approach performs worse than text-only prompting.

%% file: sec/supplementary/5_anecdotes.tex
\section{Qualitative Results}
\begin{figure*}[t]
\centering
\includegraphics[width=\textwidth]{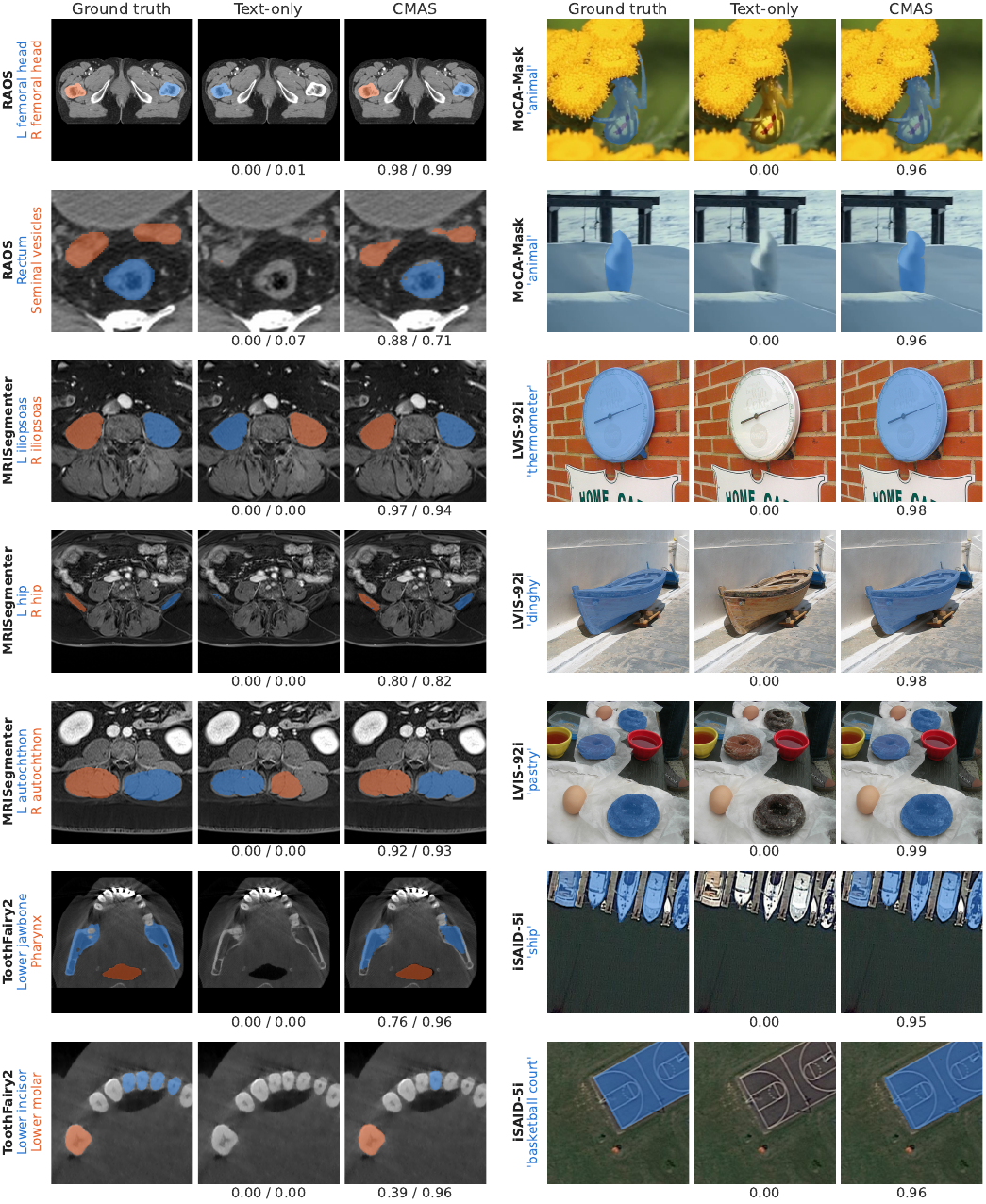}
\caption{Qualitative results: text prompt vs.\ \ours on all six datasets. Numbers: Dice (medical: slice Dice, blue / orange structure).}
\label{fig:examples}
\end{figure*}

To illustrate how cross-modal steering corrects
different segmentation failures,
\Cref{fig:examples} compares text-only prompting
with \ours across all six datasets.

\para{Medical datasets}
Text-only prompting sometimes confuses left and right
structures, such as the femoral heads on RAOS
and the iliopsoas and autochthonous muscles on
MRISegmenter, yielding Dice scores of 0.
Other structures are missed entirely, including
the hips on MRISegmenter, the jawbone and pharynx
on ToothFairy2, and the rectum and seminal vesicles
on RAOS.
\ours corrects both types of errors, achieving
Dice scores of 0.71--0.99 in most examples.
The lower incisor on ToothFairy2 remains more
challenging (0.39 Dice).

\para{Natural images}
On the illustrated examples from MoCA-Mask,
LVIS-92i, and iSAID-5i, text-only prompting
fails to detect the target objects.
In contrast, \ours recovers the requested structures,
achieving Dice scores of 0.95--0.99.

%% file: sec/supplementary/6_cost.tex
\section{Computational Cost}
\label{sup:cost}

\ours adapts segmentation models without updating
their weights, requiring substantially less compute
than parameter-efficient fine-tuning.
We quantify these savings by comparing its computational
requirements against few-shot adaptation baselines.

\Cref{tab:cost} reports the one-time cost of adaptation,
excluding test-time inference, in peta floating-point
operations (PFLOPs).
We estimate FLOPs from the number of forward and
backward passes required by each method,
using PyTorch's FLOP counter.
To avoid redundant computation, we assume cached
image features wherever applicable.
Since \ours keeps the image encoder frozen,
its features can be reused across adaptation steps.

Under this setting, \ours requires 0.47--42.2 PFLOPs
across datasets, primarily due to prompt optimization
through the fusion encoder and decoder.
This is substantially less than LoRA (31.8--170 PFLOPs),
with approximately 20$\times$ lower compute on RAOS
and 220$\times$ on MoCA-Mask.
LoRA incurs higher costs due to backpropagation
through the image encoder, which also prevents
caching of image features.

We further compare peak GPU memory requirements
in \cref{tab:costmem}.
\ours requires 4.1--8.4\,GiB across datasets,
close to text-only inference (approximately 4--5\,GiB),
compared to 8.1--22.3\,GiB for LoRA.
These results demonstrate that cross-modal steering
offers a computationally lightweight alternative
to parameter-efficient fine-tuning.

\begin{table}[t]
\centering
\small
\setlength{\tabcolsep}{3pt}
\resizebox{\linewidth}{!}{%
\begin{tabular}{@{}llcccccc@{}}
\toprule
Method & Grad. & RAOS & ToothFairy2 & MRISeg.
& MoCA-Mask & LVIS-92i & iSAID-5i \\
\midrule
Global / per-class threshold & -- & 0.15 & 0.07 & 0.20 & 0.03 & 2.7 & 0.16 \\
PerSAM~\cite{zhang2024persam} & -- & 0.14 & 0.07 & 0.18 & 0.03 & 2.5 & 0.14 \\
Geometry embedding & F, D & 17.0 & 7.7 & 25.6 & 4.7 & 435 & 23.4 \\
FS-CPL~\cite{venkataramani2026fscpl} & F, D & 20.8 & 10.0 & 33.0 & 5.8 & 528 & 29.1 \\
M2C~\cite{zhou2026m2c} & T, F, D & 1.9 & 0.93 & 3.0 & 0.40 & 35.1 & 2.0 \\
PrAda~\cite{rosi2026prada} & P & 0.34 & 0.14 & 0.59 & 0.03 & 20.2 & 0.20 \\
\midrule
LoRA~\cite{hu2022lora} & all & 31.8 & 35.9 & 37.9 & 104 & 170 & 111 \\
\midrule
Intra-modal steering & F, D & 1.6 & 0.77 & 2.5 & 0.45 & 42.2 & 2.3 \\
\textbf{\ours} & G, F, D & 1.6 & 0.78 & 2.5 & 0.47 & 42.2 & 2.3 \\
\bottomrule
\end{tabular}}
\caption{Computational cost of adaptation in $10^{15}$ FLOPs,
with cached image features where applicable.
Grad.: modules receiving gradients
(G: geometry, T: text, F: fusion, D: decoder,
P: prototypes; all: entire model).
LVIS-92i and iSAID-5i: fold 0.}
\label{tab:cost}
\end{table}

\begin{table}[t]
\centering
\small
\setlength{\tabcolsep}{4pt}
\resizebox{\linewidth}{!}{%
\begin{tabular}{@{}lcccccc@{}}
\toprule
Method & RAOS & ToothFairy2 & MRISeg. & MoCA-Mask & LVIS-92i & iSAID-5i \\
\midrule
Global / per-class threshold & 9.1 & 11.7 & 11.9 & 4.4 & 4.2 & 4.2 \\
PerSAM~\cite{zhang2024persam} & 5.3 & 5.3 & 5.2 & 4.8 & 4.1 & 4.1 \\
Geometry embedding & 4.4 & 4.2 & 4.0 & 7.5 & 6.3 & 5.8 \\
FS-CPL~\cite{venkataramani2026fscpl} & 4.4 & 4.2 & 4.0 & 7.5 & 6.3 & 5.8 \\
M2C~\cite{zhou2026m2c} & 5.3 & 5.3 & 5.3 & 10.9 & 13.5 & 14.0 \\
PrAda~\cite{rosi2026prada} & 5.7 & 5.3 & 5.2 & 4.1 & 8.1 & 4.1 \\
\midrule
LoRA~\cite{hu2022lora} & 8.1 & 9.2 & 8.7 & 22.3 & 22.3 & 22.3 \\
\midrule
Intra-modal steering & 5.3 & 4.6 & 4.1 & 8.4 & 6.8 & 6.5 \\
\textbf{\ours (ours)} & 5.3 & 4.6 & 4.1 & 8.4 & 6.8 & 6.5 \\
\bottomrule
\end{tabular}}
\caption{Peak GPU memory (GiB) of building each adaptation. Rows as in \cref{tab:cost}.}
\label{tab:costmem}
\end{table}